\documentclass[a4paper, journal]{IEEEtran}
\usepackage{color}
\usepackage{amsmath,amsfonts}
\usepackage{algorithm}
\usepackage{algorithmicx}
\usepackage{algpseudocode}
\usepackage{array}
\usepackage{cleveref}
\usepackage[caption=false,font=footnotesize,labelfont=rm,textfont=rm]{subfig}
\usepackage{textcomp}
\usepackage{stfloats}
\usepackage{url}
\usepackage{cite}
\usepackage{verbatim}
\usepackage{graphicx}
\usepackage{caption}
\usepackage{xcolor}
\usepackage{multirow}
\usepackage{multicol}
\usepackage{makecell}
\usepackage{booktabs}
\usepackage{threeparttable}
\def\BibTeX{{\rm B\kern-.05em{\sc i\kern-.025em b}\kern-.08em
    T\kern-.l_{v}67em\lower.7ex\hbox{E}\kern-.125emX}}
\usepackage{balance}

\begin{document}
\title{VideoQA-SC: Adaptive Semantic Communication for Video Question Answering}
\author{Jiangyuan Guo, Wei Chen,~\IEEEmembership{Senior Member,~IEEE}, Yuxuan Sun,~\IEEEmembership{Member,~IEEE} \\Jialong Xu,~\IEEEmembership{Member,~IEEE} and Bo Ai,~\IEEEmembership{Fellow,~IEEE}
\thanks{
Jiangyuan Guo and Wei Chen are with State Key Laboratory of Advanced Rail Autonomous Operation, and also with School of Electronic and Information Engineering, Beijing Jiaotong University, China. (e-mail: \{jiangyuanguo, weich\}@bjtu.edu.cn)

Yuxuan Sun and Jialong Xu are with the School of Electronic and Information Engineering, Beijing Jiaotong University, Beijing 100044, China. (e-mail: \{yxsun, jialongxu\}@bjtu.edu.cn)

Bo Ai is with Frontiers Science Center for Smart High-speed Railway System, and School of Electronic and Information Engineering, Beijing Jiaotong University, Beijing, China, and also with School of Information Engineering, Zhengzhou University, Zhengzhou, China. (e-mail: boai@bjtu.edu.cn)
}}

\maketitle
\pagestyle{empty}
\thispagestyle{empty}
\renewcommand{\algorithmicrequire}{\textbf{Input:}}
\renewcommand{\algorithmicensure}{\textbf{Output:}}

\begin{abstract}
Although semantic communication (SC) has shown its potential in efficiently transmitting multimodal data such as texts, speeches and images, SC for videos has focused primarily on pixel-level reconstruction. However, these SC systems may be suboptimal for downstream intelligent tasks. Moreover, SC systems without pixel-level video reconstruction present advantages by achieving higher bandwidth efficiency and real-time performance of various intelligent tasks. The difficulty in such system design lies in the extraction of task-related compact semantic representations and their accurate delivery over noisy channels. In this paper, we propose an end-to-end SC system, named VideoQA-SC for video question answering (VideoQA) tasks. Our goal is to accomplish VideoQA tasks directly based on video semantics over noisy or fading wireless channels, bypassing the need for video reconstruction at the receiver. To this end, we develop a spatiotemporal semantic encoder for effective video semantic extraction, and a learning-based bandwidth-adaptive deep joint source-channel coding (DJSCC) scheme for efficient and robust video semantic transmission. Experiments demonstrate that VideoQA-SC outperforms traditional and advanced DJSCC-based SC systems that rely on video reconstruction at the receiver under a wide range of channel conditions and bandwidth constraints. In particular, when the signal-to-noise ratio is low, VideoQA-SC can improve the answer accuracy by 5.17\% while saving almost 99.5\% of the bandwidth at the same time, compared with the advanced DJSCC-based SC system. Our results show the great potential of SC system design for video applications.
\end{abstract}


\section{Introduction}

\IEEEPARstart With the development of artificial intelligence (AI) technology, many edge machines deploy AI models to process information for intelligent tasks\cite{edge1,edge2}, which support AI-empowered applications such as remote healthcare, autonomous driving, and the Internet of Things (IoT). Semantic communication (SC) is an emerging paradigm which aims to extract the semantic information in the source data and accomplish accurate meaning delivery, ultimately completing the intelligent tasks\cite{2023Deniz}. Thanks to the effective semantic extraction by deep neural networks, SC can achieve higher data compression ratios and faster execution of intelligent tasks compared to the traditional communication. As a result, SC is widely used in many intelligent applications that require low latency and high accuracy under limited bandwidth resources, e.g., IoT networks\cite{IoT_1,IoT_2}, intelligently connected vehicle networks\cite{Vehicular_1,Vehicular_2} and smart factories \cite{smart_1}.

In a typical SC system, the transceiver is designed as a semantic codec (semantic encoder/decoder) represented by a neural network \cite{speech_1, image_2, video2}. The semantic encoder at the transmitter needs to remove data redundancy and extract compact semantic representations based on the structural characteristics of the source data. The semantic decoder at the receiver aims to process received semantic information to obtain results according to the specific intelligent task. 

For different source data modalities (speeches, texts, images, etc.), appropriate neural network architectures are essential for semantic codecs to achieve efficient SC. Long Short-Term Memorys and Transformers can be utilized to model the sequential information for texts\cite{text_1, text_3}, and convolutional neural networks (CNNs) can be utilized to extract local information for speeches\cite{speech_1,speech_2,speech_3} and images\cite{image_1,image_2,image_3}. As a novel model architecture, diffusion-based models are also increasingly applied for SC\cite{CDDM,DM-MIMO,HPQ,generativeSC}, which can learn the original data distribution through a gradual denoising process. Diffusion-based models can serve as a plug-in module to remove channel noise\cite{CDDM,DM-MIMO} or a semantic decoder to refine the degraded signals\cite{HPQ,generativeSC} at the receiver. Furthermore, deep joint source-channel coding (DJSCC) can be integrated with SC for end-to-end (E2E) training to resist wireless noise while improving the overall performance of SC\cite{digitalsc,Sem_JSCC,Yukun}. The introduction of the encryption module can also improves the security of SC systems\cite{protect}.

Unlike SC for texts or images, video-based SC faces greater challenges due to the extra \emph{temporal correlations} presented in videos. Building on traditional video compression techniques, earlier studies \cite{video1,video2} break down video transmission into the sequential transmission of several frames using conditional coding. The current frame is modeled as the conditional distribution with respect to the adjacent reference frames. Then, frames are encoded and decoded sequentially based on the reference frames in actual transmission. Some studies \cite{video5, video6} segment frames into backgrounds and key points/segments for semantic extraction and transmission. However, these approaches still focus on pixel-level video reconstruction without deeper exploration of information in videos.

For task-oriented video transmission, it is necessary to extract deeper semantics containing the meaning of video contents. Shao \emph{et al.} \cite{video3} proposed a video transmission system TOCOM-TEM for edge video analytics tasks. The system includes task-relevant feature extraction, feature encoding and joint inference modules. To reduce the communication overhead, deterministic information bottleneck and temporal entropy models are leveraged. Nevertheless, its semantic encoding of video is still decomposed into conditional coding of sequential frames. This method lack global \textit{spatiotemporal modeling} of videos, leading to inefficient semantic extraction. Wan \emph{et al.} \cite{3dcnn} proposed to use 3D-CNNs to extract spatiotemporal correlations for human activity recognition tasks, with limitations such as high computational overheads and low operational efficiency. Both these methods directly extract high-level semantics from pixel-level videos, which is challenging and inefficient. Overall, current video-based SC research mainly focuses on video reconstruction, with few investigations and developments for other intelligent functionalities. Few existing task-oriented video modeling methods are not effective for extracting high-level video semantics.

Video question answering (VideoQA), where machines automatically answer natural language questions with video contents, is an intelligent task in the popular visual-language understanding domain. Solving VideoQA tasks enables innovative applications in human-machine interactions such as virtual reality, smart cities, and the metaverse. The proliferation of multimedia applications and the extensive deployment of cameras have led to a significant presence of videos in machines, affecting both human-machine and machine-machine communications. 

Compared with image-based visual question answering, VideoQA includes a broader range of question types. It involves not only recognition of visual objects, actions, and events, but also reasoning of spatiotemporal and causal relationships, making it more challenging \cite{VideoQA-survey}. The key of VideoQA is to understand video contents with questions, which can be decoupled into text, video analysis and fusion. On the one hand, the multi-task (MT) learning approaches for SC can provide a useful view\cite{multi-task1,multi-task2,multi-task3}, which achieve efficient communication for multiple semantic tasks with multimodal data. Feature parsing and decoding with multi-exit task heads are exploited in \cite{multi-task1} and \cite{multi-task2} for handling various semantic tasks given the same features, respectively. Zhang \emph{et al.}\cite{multi-task3} proposed a unified MT system for SC, which can handles semantic tasks involving texts and images. In general, these approaches lack effective exploration and design for videos. On the other hand, traditional VideoQA works give some methods for video processing\cite{VideoQA_1,VideoQA_2,VideoQA_3,VideoQA_4,VideoQA_5}. Some works jointly extract frame features and motion features to get effective video representations \cite{VideoQA_1,VideoQA_2,VideoQA_3}. The generic backbone like Vision Transformers are used to obtain general video representations in \cite{VideoQA_4,VideoQA_5}. However, the video features extracted by traditional VideoQA methods are usually of high dimensions, which may not meet bandwidth constraints in wireless networks. Moreover, channel fading and noise also affect the accurate transmission of video features, resulting in degradation of VideoQA performance.

Typically, the development of SC systems for VideoQA tasks encounters two key challenges:
\begin{enumerate}
    \item How to model the spatiotemporal correlations of videos to achieve efficient semantic extraction?
    \item How to mitigate the effects of wireless channel degradation and meet bandwidth constraints while maintaining the performance of VideoQA tasks?
\end{enumerate}

In this paper, we investigate an E2E SC system named VideoQA-SC for VideoQA tasks. The proposed VideoQA-SC mainly incorporates two customized modules to address the above challenges: a \textit{spatiotemporal} video semantic encoder and a \textit{learning-based bandwidth-adaptive} joint source-channel (JSC) encoder/decoder. Experimental results demonstrate that the proposed VideoQA-SC achieves noise robustness and bandwidth efficiency.

The main contributions of this paper are summarized as follows:
\begin{enumerate}
    \item \textit{An E2E SC System for VideoQA Tasks}: We propose an E2E SC system called \emph{VideoQA-SC} for VideoQA tasks. VideoQA-SC exploits the efficient video semantic extraction and the bandwidth-adaptive DJSCC transmission to fully leverage video information, which is noise robustness and bandwidth efficiency with promising task performance.
    \item \textit{Spatiotemporal Semantic Encoder}: We propose a spatiotemporal semantic encoder to extract compact and comprehensive video semantics for transmission. Transformer and the graph neural network are utilized to model the temporal and spatial correlations of videos, which is beneficial for understanding video contents.
    \item \textit{Cross-Attention Based JSC Encoder/Decoder}: We propose a dual-branch cross-attention Transformer architecture as both the JSC encoder and decoder with the learnable rate embedding shared between the transmitter and receiver. The structure allows for progressive refinement of the semantics at both the transmitter and receiver by the cross-attention mechanism.
    \item \textit{Learning-Based Adaptive Bandwidth Allocation}: We develop a series of learning-based rate predictors to allocate bandwidth to video semantics for transmission. The rate predictors can learn the importance of different tokens in semantics, improving the bandwidth efficiency of SC systems. Moreover, the rate predictors allow other useful information, e.g., channel state information, to serve as additional guidance of bandwidth allocation, demonstrating good scalability for learning-based bandwidth allocation methods.
    \item \textit{Experimental Analysis}: We verify the performance of VideoQA-SC on the TGIF-QA\cite{TGIF-QA} dataset. Experiments demonstrate that VideoQA-SC outperforms traditional communication systems and other DJSCC-based SC systems under a wide range of channel conditions and bandwidth constraints. In particular, VideoQA-SC improves $5.17\%$ VideoQA accuracy while achieving nearly $99.5\%$ bandwidth savings compared with the DJSCC-based SC system over the additive Gaussian white noise (AWGN) channel at $0$ dB signal-to-noise ratio (SNR).
\end{enumerate}

\begin{figure}[!t]
\centering
\includegraphics[width=0.42\textwidth]{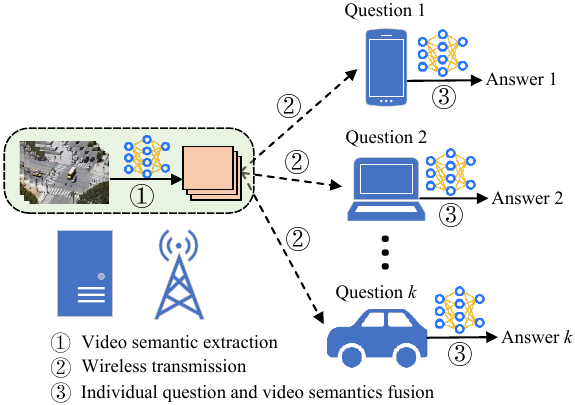}
\caption{An application scenario for VideoQA-SC.}
\label{Fig:VideoQA}
\end{figure}
The rest of this paper is organized as follows. The system model and the process for performing VideoQA tasks are introduced in \Cref{Section:system model}. We explain our proposed methods and the detailed network architectures in \Cref{Section:The Proposed Method}. \Cref{Section:Experiments} provides the quantified experimental results and the comparison with existing advanced methods. Finally, \Cref{Section:Conclusion} summarizes this paper and gives conclusions.

\textit{Notations}: In this paper, lowercase letters, e.g., ${x}$, denote scalars. Bold lowercase letters, e.g., $\mathbf{x}$, denote vectors and bold uppercase letters, e.g., $\mathbf{X}$, denote matrices or tensors. $\mathbf{I}$ denotes the identity matrix. $\mathcal{CN}(\mu,\sigma^{2})$ and $\mathcal{N}(\mu,\sigma^{2})$ denote the complex Gaussian distribution and the real Gaussian distribution with mean $\mu$ and covariance $\sigma^{2}$, respectively. $\log_{2}$ denotes the logarithm to base $2$ and $\log$ denotes the natural logarithm. $(\cdot)^{T}$ denotes the transpose and $(\cdot)^{*}$ denotes the conjugate transpose. $\mathbb{R}$ and $\mathbb{C}$ denote the real set and the complex set, respectively. $\mathbb{E}[\cdot]$ denotes the statistical expectation operation. $\operatorname{Uniform}(a,b)$ denotes the uniform distribution with start $a$ and end $b$.

\section{System Model}\label{Section:system model}
In this section, we introduce the VideoQA-SC workflow to perform VideoQA tasks and establish an optimization model for the entire system with bandwidth constraints.

Fig. \ref{Fig:VideoQA} shows an application scenario for VideoQA-SC. There are many terminal devices simultaneously requesting access to the same surveillance videos with different questions. The transmitter, e.g., edge server, extracts video semantics containing comprehensive video contents and sends them to all terminal devices. Then, each terminal device independently completes VideoQA to predict its own answer.

Our work focuses on the multi-choice VideoQA tasks. Given the video $\mathbf{X}_{v}\in \mathbb{R}^{l_{v}\times{3}\times{x}\times{y}}$ with $l_{v}$ frames and the question $\mathbf{X}_{q}\in \mathbb{R}^{l_{q}\times{d}_{q}}$ with $l_{q}$ tokens, VideoQA aims to predict an answer $a$ by exploiting both video and text information:
\begin{equation}
a^{\star} = \mathop{\arg\max}\limits_{a\in{\mathcal{A}}}o_{\boldsymbol{\omega}}(a|\mathbf{X}_{q},\mathbf{X}_{v},\mathcal{A}),
\end{equation}
where $a^{\star}$ is the predicted answer chosen from the candidate answers, i.e., multiple choices or a predefined global answer set, denoted as $\mathcal{A}$, and $o_{\boldsymbol{\omega}}(\cdot)$ is the VideoQA model with the learnable parametric vector $\boldsymbol{\omega}$. 

\begin{figure*}[!t]
\centering
\includegraphics[width=0.9\textwidth]{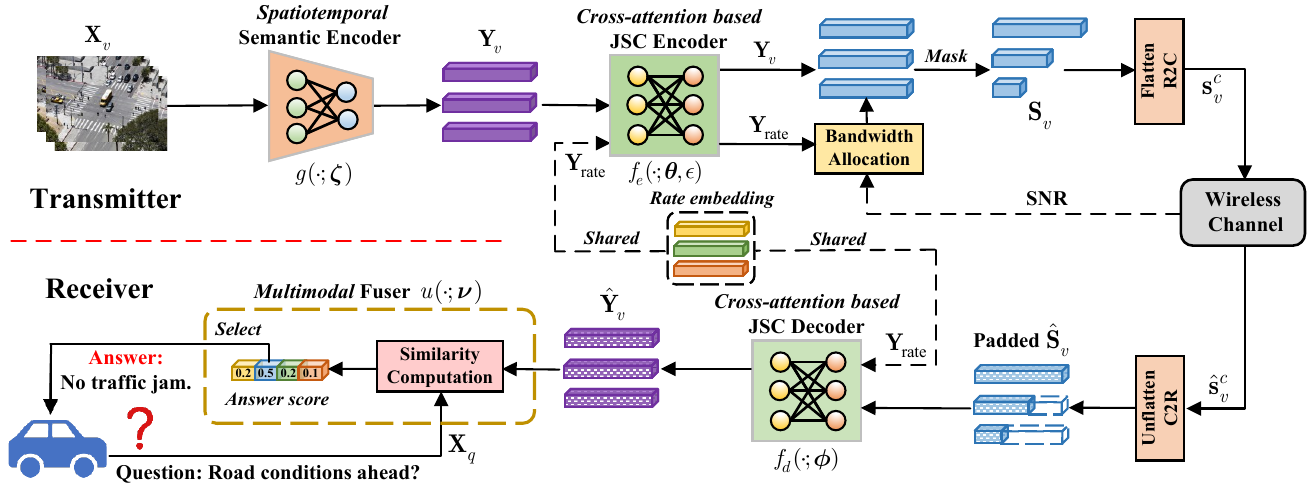}
\caption{Overview of the proposed VideoQA-SC.}
\label{Fig:overview}
\end{figure*}

As illustrated in Fig. \ref{Fig:overview}, the whole process of VideoQA-SC mainly includes 3 parts:
\subsubsection{Transmitter}
The transmitter first extracts the low-dimensional semantics $\mathbf{Y}_{v}\in \mathbb{R}^{l_{v}\times{d}}$ from the input video $\mathbf{X}_{v}$ using the spatiotemporal semantic encoder $g(\cdot;\boldsymbol{\zeta})$ with the learnable parametric vector $\boldsymbol{\zeta}$. Then, the JSC encoder with rate predictors $f_{e}(\cdot;\boldsymbol{\theta},\boldsymbol{\epsilon})$ processes $\mathbf{Y}_{v}$ into $\mathbf{S}_{v}\in \mathbb{R}^{l_{v}\times{d}}$, part of whose channels are masked as zero. $\boldsymbol{\theta}$ and $\boldsymbol{\epsilon}$ are the learnable parametric vectors of the JSC encoder and rate predictors, respectively. Non-zero channels of $\mathbf{S}_{v}$ are flattened into continuous-valued real symbols. Finally, complex channel input symbols $\mathbf{s}_{v}^{c}\in \mathbb{C}^{n}$ are obtained by converting each two real symbols into one complex symbol. The process of $\mathbf{X}_{v}$ at the transmitter can be expressed as:
\begin{equation}
    \mathbf{s}_{v}^{c}=\operatorname{R2C}(\operatorname{Flatten}(f_{e}(g(\mathbf{X}_{v};\boldsymbol{\zeta});\boldsymbol{\theta},\boldsymbol{\epsilon}))),
\end{equation}where $\operatorname{R2C}(\cdot)$ and $\operatorname{Flatten}(\cdot)$ are the real-to-complex and flattening operations, respectively.
$\frac{1}{n}\mathbf{s}_{v}^{c}(\mathbf{s}_{v}^{c})^{*}\le{1}$ is imposed to satisfy the average power constraint at the transmitter.

Here, we define $R=\frac{n}{l_{v}\times{3}\times{x}\times{y}}\le1$ as the \textit{bandwidth compression ratio} (BCR), which represents the average length of channel input symbols encoded for each source symbol.

\subsubsection{Channel}
The encoded channel input symbols $\mathbf{s}_{v}^{c}$ are transmitted over the noisy wireless channel. For the AWGN channel, the received symbols can be expressed as:
\begin{equation}
\mathbf{\hat{s}}_{v}^{c}=\mathbf{s}_{v}^{c}+\mathbf{n},
\end{equation}
where $\mathbf{n}\in \mathbb{C}^{n}$ consists of independent and identically distributed (i.i.d.) samples following $\mathcal{CN}(0,\sigma^{2}_{n}\mathbf{I})$. $\sigma^{2}_{n}$ denotes the average noise power. For Rayleigh block fading channels, an additional channel gain $h\in\mathbb{C}^{1}$ is introduced for each $\mathbf{s}_{v}^{c}$:
\begin{equation}
\mathbf{\hat{s}}_{v}^{c}=h\mathbf{s}_{v}^{c}+\mathbf{n},
\end{equation}
where $h$ is sampled from the Rayleigh distribution $\operatorname{Rayleigh}(\sigma_{h})$ with fading parameter $\sigma_{h}$. Here, the statistical SNR and the current SNR for each $\mathbf{s}_{v}^{c}$ are given by $\frac{h^{2}}{\sigma_{n}^{2}}$ and $\frac{\sigma_{h}^{2}}{\sigma_{n}^{2}}$, respectively.
\subsubsection{Receiver}
The received complex channel output symbols $\mathbf{\hat{s}}_{v}^{c}$ are first converted to real symbols, which are further unflattened and padded with zeros to form $\mathbf{\hat{S}}_{v}\in \mathbb{R}^{l_{v}\times{d}}$. The receiver then decodes the video semantics $\mathbf{\hat{Y}}_{v}\in \mathbb{R}^{l_{v}\times{d}}$ from $\mathbf{\hat{S}}_{v}$ by the JSC decoder $f_{d}(\cdot;\boldsymbol{\phi})$ with the learnable parametric vector $\boldsymbol{\phi}$. Subsequently, the receiver utilizes the multimodal fuser $u(\cdot;\boldsymbol{\nu})$ with the learnable parametric vector $\boldsymbol{\nu}$ to interact with video and text information and predicts the corresponding answer $a^\star$ to the question $\mathbf{X}_{q}$. The process of answer prediction at the receiver can be expressed as:
\begin{equation}
a^{\star}=u(f_{d}(\operatorname{Pad}(\operatorname{Unflatten}(\operatorname{C2R}(\mathbf{\hat{s}}_{v}^{c})));\boldsymbol{\phi}),\mathbf{X}_{q};\boldsymbol{\nu}),
\end{equation}
where $\operatorname{Pad}(\cdot)$, $\operatorname{Unflatten}(\cdot)$ and $\operatorname{C2R}(\cdot)$ denote zero-padding, unflattening and complex-to-real operations, respectively.

The goal of VideoQA-SC is to maximize the average accuracy of VideoQA on testing data for a given bandwidth $B$ by optimizing all learnable parametric vectors $\boldsymbol{\zeta}$, $\boldsymbol{\theta}$, $\boldsymbol{\epsilon}$, $\boldsymbol{\phi}$ and $\boldsymbol{\nu}$, which can be formulated as:
\begin{align}
\begin{aligned}
\max_{\boldsymbol{\zeta},\boldsymbol{\theta},\boldsymbol{\epsilon},\boldsymbol{\phi},\boldsymbol{\nu}}&\quad\mathbb{E}[\operatorname{ACC}(a^{\star}, a_\text{label})] \\
\mathrm{s.t.}&\quad R\le{B},
\end{aligned}
\end{align}
where $\operatorname{ACC}(\cdot,\cdot)$ is an indicator function (1 only if $a^{\star}=a_\text{label}$ and 0 otherwise).

\section{The Proposed Method}\label{Section:The Proposed Method}
In this section, we describe the proposed VideoQA-SC system in detail, which includes the spatiotemporal semantic encoder, cross-attention based JSC encoder/decoder and learning-based adaptive bandwidth allocation. Then, we introduce the training strategy of VideoQA-SC.
\subsection{Spatiotemporal Semantic Encoder}
We develop a spatiotemporal semantic encoder $g(\cdot;\boldsymbol{\zeta})$ to extract the video semantics by modeling the spatial and temporal correlations of videos. The purpose of $g(\cdot;\boldsymbol{\zeta})$ is to extract the coarse-grained semantics that is beneficial to fully understand the video content. In this way, although there may be multiple receivers with different inquiries, they are able to discern the video content pertinent to their specific questions through the processed video semantics, enabling them to perform their own analysis without video recovery.

Given that consecutive frames in a video typically have identical backgrounds, substantial spatial and temporal redundancies exist, necessitating removal to enhance semantic extraction. Inspired by VideoQA works that extracted frame features and motion features to get video representations \cite{VideoQA_1,VideoQA_2,VideoQA_3}, the proposed spatiotemporal semantic encoder $g(\cdot;\boldsymbol{\zeta})$ operates mainly on object-level and frame-level features to capture the changes of visual objects while removing redundancy, thereby extracting compact and comprehensive semantic representations. 

To reduce the use of computational and transmission resources, we initially apply uniform interval sparse sampling across each video, choosing $l_{v}$ frames as keyframes to make up $\mathbf{X}_{v}$. The $l_{v}$ keyframes are divided into $l_{c}$ clips, each with a length of $l_{f}$ frames. For simplicity, we assume that each clip operates independently, and $g(\cdot;\boldsymbol{\zeta})$ is designed to only capture the correlations within the frames of a single clip and produce semantics for the clip.

\begin{figure*}[!t]
\centering
\includegraphics[width=0.80\textwidth]{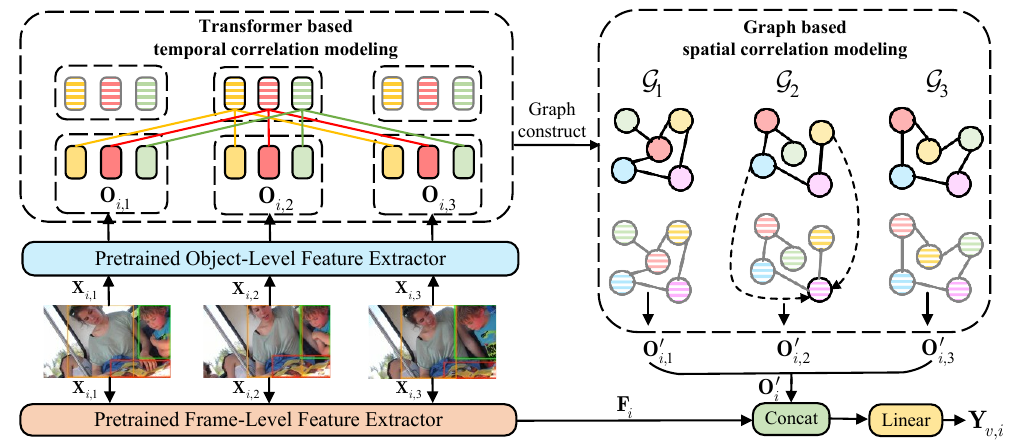}
\caption{Temporal and spatial modeling of the $i$-th video clip. Shapes with the same color represent the same objects in different frames. Striped shapes indicate fused features after corresponding modeling. An object-level feature extractor and a frame-level feature extractor are used to preprocess all frames in the $i$-th clip for further processing.}
\label{Fig:TS_modeling}
\end{figure*}

The process of temporal and spatial modeling is illustrated in Fig. \ref{Fig:TS_modeling}. 
For the $i$-th clip $\mathcal{X}_{i}=\left\{\mathbf{X}_{i,1},\mathbf{X}_{i,2}, 
\cdots, \mathbf{X}_{i,l_{f}}\right\}$, $\mathbf{X}_{i,j}\in\mathbb{R}^{3\times{x}\times{y}}$ represents the $j$-th frame in the clip. We first use a pretrained object detector from \cite{Anderson_2018_CVPR} and pretrained standard ResNet-101 from \cite{He_2016_CVPR} without classification heads as the preprocessing networks to process all $\mathbf{X}_{i,j}$ $(j\in{\left [1,l_{f} \right ]})$, obtaining the object-level feature $\mathbf{O}_{i,j}\in{\mathbb{R}^{r\times{m}}}$ and frame-level feature $\mathbf{F}_{i,j}\in{\mathbb{R}^{m}}$, respectively. Actually, any network which is able to extract object-level or frame-level features can serve as the preprocessing network. Let $r$ be the number of detected objects in $\mathbf{X}_{i,j}$ and $m$ be the channel dimension of both the object-level features and the frame-level features. We concatenate all $\mathbf{O}_{i,j}$ and $\mathbf{F}_{i,j}$ to get the $i$-th clip object-level feature $\mathbf{O}_{i}\in{\mathbb{R}^{l_{f}\times{r}\times{m}}}$ and the $i$-th clip frame-level feature $\mathbf{F}_{i}\in{\mathbb{R}^{l_{f}\times{m}}}$, respectively. 

In the Transformer architecture\cite{Attention}, the self-attention mechanism utilizes the learnable matrix $\mathbf{W}^{QKV}$ to generate the query ($\mathbf{Q}$), key ($\mathbf{K}$), and value ($\mathbf{V}$) representations from inputs. Then, the attention is given by the scaled dot-product of $\mathbf{Q}$ and $\mathbf{K}$ with softmax normalization to achieve weighted information aggregation, which can effectively capture dependencies between different words in the sequence. Inspired by this, we input each $\mathbf{O}_{i}$ into stacked Transformer blocks to facilitate the interaction of the same object features in different frames in the $i$-th clip. Here, the number of frames in the clip corresponds to the sequence length in the original Transformer. By using self-attention mechanism, we can obtain an aggregated representation of each object $\mathbf{O}_{i,j,k}\in\mathbb{R}^{m}$ $(k\in\left [1,r \right ])$ to learn the object-level temporal correlations. The aggregation can be expressed as:
\begin{equation}
    \sum_{p=1}^{l_{f}}\alpha_{i,k,j,p}\mathbf{O}_{i,p,k}\to \mathbf{O}_{i,j,k},
\label{eq_attention}
\end{equation}
where $\alpha_{i,k,j,p}$ is the attention score. Eq. (\ref{eq_attention}) denotes each object $\mathbf{O}_{i,j,k}$ is the fusion of the same object in different frames ($\mathbf{O}_{i,p,k}, p\in\left[1,l_{f} \right]$) along the time dimension.

Subsequently, following the work \cite{VideoQA_3}, we construct a graph $\mathcal{G}_{i}$ for every $\mathbf{O}_{i,j}$, where each node and each edge in $\mathcal{G}_{i}$ denotes one object ($\mathbf{O}_{i,j,k}$) and the spatial correlations between two objects, respectively. Based on the constructed graph, we perform graph convolution operations on every $\mathbf{O}_{i,j}$, exploiting the spatial information between different objects in one frame to utilize spatial interaction. Finally, after average pooling along the object dimension, the processed object-level feature $\mathbf{O'}_{i}\in\mathbb{R}^{l_{f}\times{m}}$ is concatenated with the frame-level feature $\mathbf{F}_{i}$. The concatenated feature is fed into a linear layer to map to the video semantics $\mathbf{Y}_{v,i}\in\mathbb{R}^{l_{f}\times{d}}$. All clip-level representations $\mathbf{Y}_{v,i}$ are concatenated to obtain the entire video semantics $\mathbf{Y}_{v}\in\mathbb{R}^{l_{v}\times{d}} $.

\begin{figure}[!t]
\centering
\includegraphics[width=0.4\textwidth]{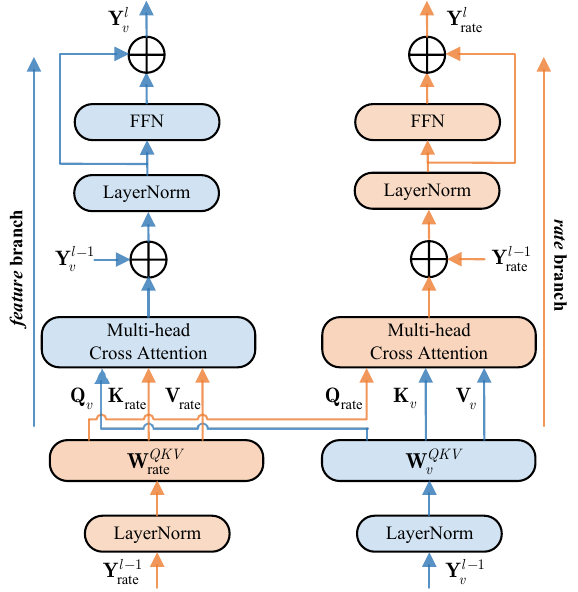}
\caption{The structure of the dual-branch cross-attention Transformer block in the JSC encoder/decoder.}
\label{Fig:cross_attention}
\end{figure}

\subsection{Cross-Attention Based DJSCC Transmission}\label{Section:CA}
We apply DJSCC technology to VideoQA-SC and design a symmetric JSC encoder/decoder to overcome wireless channel degradation, enabling accurate transmission of video semantics $\mathbf{Y}_{v}$. The learnable embedding shared between the transmitter and receiver is developed, which progressively refines $\mathbf{Y}_{v}$ in the form of cross-attention during the JSC encoding and decoding processes.

We use Transformer to construct the backbone of both the JSC encoder and decoder. In DJSCC, code rate guidance allows the refinement of the latent representations, thereby generating channel input symbols adaptive to bandwidth constraints, which motivates us to provide code rate guidance to the encoding and decoding processes of video semantics. 

Different from directly using the human-defined code rate, we design the code rate as learnable parameters and interact with the video semantics in the form of cross attention. Specifically, we introduce a learnable parametric tensor, with the same shape as $\mathbf{Y}_{v}$, termed rate embedding $\mathbf{Y}_\text{rate}$, in the JSC encoder $f_{e}(\cdot;\boldsymbol{\theta}, \boldsymbol{\epsilon})$ and the JSC decoder $f_{d}(\cdot;\boldsymbol{\phi})$. The encoding and decoding processes of $\mathbf{Y}_{v}$ are both guided by $\mathbf{Y}_\text{rate}$ shared between the transmitter and receiver. $\mathbf{Y}_\text{rate}$ allocates different attention to $\mathbf{Y}_{v}$ under different bandwidth constraints to aid encoding and decoding processes. Different from using random initialization, we set the initial values of $\mathbf{Y}_\text{rate}$ as a set of learnable parameters. By training under different bandwidth constraints, the model learns different initial values for $\mathbf{Y}_\text{rate}$. Each set of initial values can be viewed as prior information, representing the initial attention allocation results for semantics of different
frames under a specific bandwidth constraint. During testing, for a specific bandwidth constraint, the corresponding initial values are selected to initialize $\mathbf{Y}_\text{rate}$ for inference. Furthermore, $\mathbf{Y}_\text{rate}$ can aid the JSC encoder in achieving variable-length coding of $\mathbf{Y}_{v}$, which will be explained in detail in \Cref{Section:bandwidth allocation}. 

Since the JSC encoder and decoder have similar network structures, we take the process of the JSC encoder as an example to describe the interaction between $\mathbf{Y}_\text{rate}$ and $\mathbf{Y}_{v}$. As illustrated in Fig. \ref{Fig:cross_attention}, the proposed cross-attention Transformer block consists of two symmetric branches (\textit{rate} branch and \textit{feature} branch) to process $\mathbf{Y}_\text{rate}$ and $\mathbf{Y}_{v}$, respectively. Each branch is a standard Transformer block in Vision Transformer. 

Starting from the projection of two embeddings,$\mathbf{Y}_\text{rate}\in{\mathbb{R}^{l_{v}\times{d}}}$ is transformed into $\mathbf{Q}_\text{rate}\in\mathbb{R}^{l_{v}\times{d}}$, $\mathbf{K}_\text{rate}\in\mathbb{R}^{l_{v}\times{d}}$ and  $\mathbf{V}_\text{rate}\in\mathbb{R}^{l_{v}\times{d}}$ by $\mathbf{W}^{QKV}_\text{rate}$, and $\mathbf{Y}_{v}\in{\mathbb{R}^{l_{v}\times{d}}}$ is transformed into $\mathbf{Q}_{v}\in\mathbb{R}^{l_{v}\times{d}}$, $\mathbf{K}_{v}\in\mathbb{R}^{l_{v}\times{d}}$ and $\mathbf{V}_{v}\in\mathbb{R}^{l_{v}\times{d}}$ by $\mathbf{W}^{QKV}_{v}$. The proposed cross-attention mechanism utilizes the scaled dot-product of $\mathbf{Q}_\text{rate}$ and $\mathbf{K}_{v}$ to generate attention for the \textit{rate} branch and the scaled dot-product of $\mathbf{Q}_{v}$ and $\mathbf{K}_\text{rate}$ to generate attention for the \textit{feature} branch. The implementation of cross-attention for $\mathbf{Y}_\text{rate}$ to $\mathbf{Y}_{v}$ and $\mathbf{Y}_{v}$ to $\mathbf{Y}_\text{rate}$ can be formulated as:
\begin{equation}
\operatorname{CA}(\mathbf{Y}_\text{rate},\mathbf{Y}_{v})=\operatorname{softmax}\left(\frac{\mathbf{Q}_\text{rate} \mathbf{K}^{T}_{v}}{\sqrt{d} }\right)\mathbf{V}_{v},
\end{equation}
and
\begin{equation}
\operatorname{CA}(\mathbf{Y}_{v},\mathbf{Y}_\text{rate})=\operatorname{softmax}\left(\frac{\mathbf{Q}_{v} \mathbf{K}^{T}_\text{rate}}{\sqrt{d} }\right)\mathbf{V}_\text{rate},
\end{equation}
respectively. 

Then, the process of the entire Transformer block for the \textit{rate} branch and the \textit{feature} branch can be formulated as:

\begin{align}
\begin{aligned}
\tilde{\mathbf{Y}} _\text{rate}^{l} &= \operatorname{MHCA}(\operatorname{LN}(\mathbf{Y}_\text{rate}^{l-1}),\operatorname{LN}(\mathbf{Y}_{v}^{l-1})) + \mathbf{Y} _\text{rate}^{l-1}, \\
\mathbf{Y}_\text{rate}^{l} &= \operatorname{FFN}(\operatorname{LN}(\tilde{\mathbf{Y}}_\text{rate}^{l}))+\operatorname{LN}(\tilde{\mathbf{Y}}_\text{rate}^{l}) ,
\end{aligned}
\label{eq_1}
\end{align}
and
\begin{align}
\begin{aligned}
\tilde{\mathbf{Y}} _{v}^{l} &= \operatorname{MHCA}(\operatorname{LN}(\mathbf{Y}_{v}^{l-1}),\operatorname{LN}(\mathbf{Y}_\text{rate}^{l-1})) + \mathbf{Y} _{v}^{l-1}, \\
\mathbf{Y}_{v}^{l} &= \operatorname{FFN}(\operatorname{LN}(\tilde{\mathbf{Y}}_{v}^{l}))+\operatorname{LN}(\tilde{\mathbf{Y}}_{v}^{l}),
\end{aligned}
\label{eq_2}
\end{align}
respectively. In Eq. (\ref{eq_1}) and (\ref{eq_2}), $\mathbf{Y}_\text{rate}^{l-1}$ and $\mathbf{Y}_{v}^{l-1}$ denote the inputs of the $l$-th Transformer block of the two branches, and $\mathbf{Y}_\text{rate}^{l}$ and $\mathbf{Y}_{v}^{l}$ denote the outputs of the $l$-th Transformer blocks of the two branches. $\operatorname{MHCA}(\cdot)$ represents $\operatorname{CA}(\cdot)$ function with multi-heads. $\operatorname{LN}(\cdot)$ represents the layer normalization in the Transformer. $\operatorname{FFN}(\cdot)$ represents two linear layers with $\operatorname{GeLU}(\cdot)$ activation function. 

The proposed symmetric dual-branch cross-attention Transformer block allows two types of embedding ($\mathbf{Y}_\text{rate}$ and $\mathbf{Y}_{v}$) to interact information in the form of cross-attention, thereby promoting information flow across both branches. As $\mathbf{Y}_{v}^{l-1}$ is updated to $\mathbf{Y}_{v}^{l}$, $\mathbf{Y}_\text{rate}^{l-1}$ is also updated to $\mathbf{Y}_\text{rate}^{l}$, which provides dynamic rate guidance to scale each feature in $\mathbf{Y}_{v}^{l}$ in the next Transformer block. During the interaction of the two branches, $\mathbf{Y}_\text{rate}$ and $\mathbf{Y}_{v}$ refine each other iteratively and finally contribute to the generation of real symbols $\mathbf{S}_{v}$. 

After flattening and the real-to-complex operation, the channel input symbols $\mathbf{s}_{v}^{c}$ are transmitted through the noisy channel. The JSC decoder $f_{d}(\cdot;\boldsymbol{\phi})$ has the same structure as the JSC encoder, consisting of stacked dual-branch cross-attention Transformer blocks. The JSC decoder exploits the same rate embedding $\mathbf{Y}_\text{rate}$ and progressively decodes the video semantics $\hat{\mathbf{Y}}_{v}$ based on the noisy channel output symbols $\hat{\mathbf{s}}_{v}$ at the receiver.

\begin{figure*}[!t]
\centering
\includegraphics[width=0.90\textwidth]{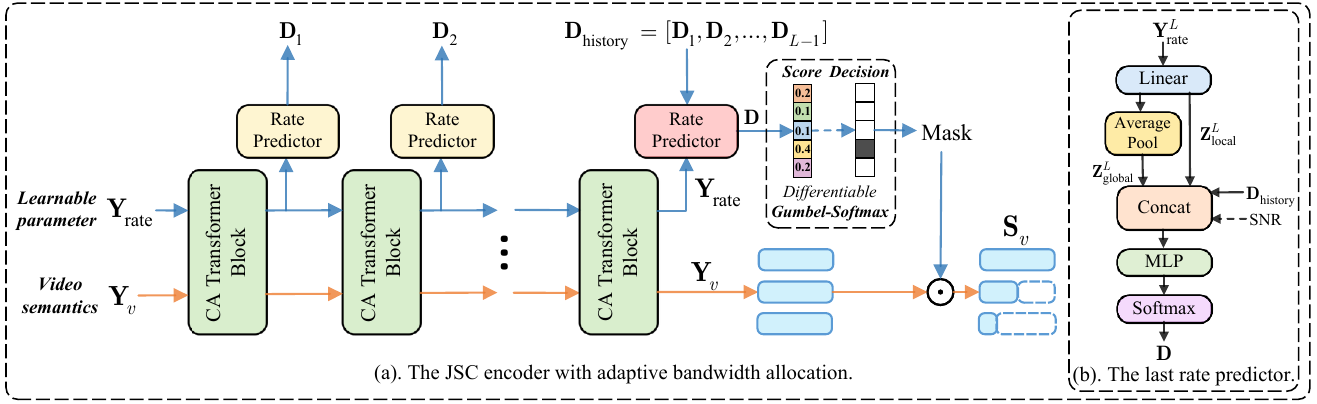}
\caption{The JSC encoder with adaptive bandwidth allocation. CA Transformer block denotes the proposed dual-branch cross-attention Transformer block in \Cref{Section:CA}.}
\label{Fig:JSC}
\end{figure*}
\subsection{Learning-Based Adaptive Bandwidth Allocation} \label{Section:bandwidth allocation}
We propose a learning-based adaptive bandwidth allocation approach to generate channel input symbols of variable lengths, further improving the bandwidth efficiency of VideoQA-SC.

For the full use of limited bandwidth resources, flexible bandwidth allocation is required, e.g., more bandwidth for important information and less bandwidth for less important information. Statistical-based methods and learning-based methods can both be employed to measure the importance of features. Statistical-based methods, such as the feature entropy estimation \cite{video2}, explicitly model the importance of features based on their entropy. Learning-based methods, e.g., distinguishing informative features and uninformative features with scaling factors in Batch Normalization layers \cite{BN}, implicitly model the importance of features.
    
According to the cross-attention mechanism, $\mathbf{Y}_\text{rate}$ described in \Cref{Section:CA} can be seen as the score metric that dynamically scales elements of $\mathbf{Y}_{v}$, causing us to measure the importance of features based on $\mathbf{Y}_\text{rate}$. Our learning-based adaptive bandwidth allocation approach exploits $\mathbf{Y}_\text{rate}$ as guidance, which sparsifies the channels of each token output by the JSC encoder to generate channel input symbols of variable lengths. Specifically, given the output of the last dual-branch Transformer block $\mathbf{Y}_\text{rate}\in{\mathbb{R}^{l_{v}\times d}}$ and $\mathbf{Y}_{v}\in{\mathbb{R}^{l_{v}\times d}}$, we develop a series of rate predictors parameterized by $\boldsymbol{\epsilon}$ to predict the retained dimension for each token $\mathbf{Y}_{v,i}\in{\mathbb{R}^{d}}$ by $\mathbf{Y}_\text{rate}$. Then, a binary mask matrix $\mathbf{M}\in{\mathbb{R}^{l_{v}\times d}}$ is generated for channel masking, with the $i$-th row containing the first $k_{i}$ ones followed by $(d-k_{i})$ zeros ($k_{i}\in{\left[0,d\right]}$). $\mathbf{M}$ is used to retain the first $k_{i}$ channels of $\mathbf{Y}_{v,i}$ and mask the rest channels: 
\begin{equation}
    \mathbf{Y}_{v}\odot\mathbf{M}\to\mathbf{Y}_{v},
\end{equation}
where $\odot$ is Hadamard product. 

To facilitate the learning of bandwidth allocation by neural networks, we set up $q=\log_{2} d$ fixed candidate bandwidth for each token. In other words, for each $\mathbf{Y}_{v,i}$, $k_{i}$ is selected from these $q$ fixed values ($k_{i}\in\left \{ 2,4,8,\cdots,d \right \} $). Therefore, the channel sparsification problem can be seen as a classification problem in selecting the most suitable category from $q$ categories for each token $\mathbf{Y}_{v,i}$. However, the output of the rate predictors is the probability of $q$ categories, which need to sample a specific ``class'' for each token. In E2E training, this sampling operation is non-differentiable, making it impossible to update the parameters of rate predictors through gradient descent. To overcome the problem, we employ the classical Gumbel-Softmax \cite{gumbel_softmax} trick to implement differentiable sampling operations. Next, we will elaborate on the process of channel masking.

As illustrated in Fig. \ref{Fig:JSC}, after every cross-attention Transformer block, a rate predictor is introduced to give a rate prediction for the current feature $\mathbf{Y}_{v}$. Consider that we have $L$ cross-attention Transformer blocks and $L$ rate predictors. The $l$-th ($l\in\left [1,L-1\right ]$) rate predictor takes the current rate embedding $\mathbf{Y}_\text{rate}^{l}$ as input. First, the $l$-th rate predictor projects $\mathbf{Y}_\text{rate}^{l}$ using a linear layer to model its local information:
\begin{equation}
    \mathbf{Z}^{l}_\text{local} = \operatorname{Linear}(\mathbf{Y}_\text{rate}^{l})\in{\mathbb{R}^{l_{v}\times d}},
\end{equation}
where $\operatorname{Linear}(\cdot)$ denotes the linear layer with $\operatorname{GeLU}(\cdot)$ activation function. We apply average pooling to $\mathbf{Z}^{l}_\text{local}$ along the token dimension to obtain the global information:
\begin{equation}
    \mathbf{Z}^{l}_\text{global} = \operatorname{AveragePool}(\mathbf{Z}_\text{local}^{l})\in{\mathbb{R}^{1\times{d}}},
\end{equation}
where $\operatorname{AveragePool}(\cdot)$ denotes the average pooling operation. Then, $\mathbf{Z}^{l}_\text{global}$ is expanded from $\mathbb{R}^{1\times{d}}$ to $\mathbb{R}^{{l_v}\times{d}}$. After that, the rate predictor combines the local and global information along the channel dimension:
\begin{equation}
    \mathbf{Z}_\text{rate}^{l} = \operatorname{concat}(\mathbf{Z}^{l}_\text{local},\mathbf{Z}^{l}_\text{global})\in{\mathbb{R}^{l_{v}\times{2d}}},
\end{equation}
where $\operatorname{concat(\cdot)}$ denotes the concatenation along the channel dimension.
Then, $\mathbf{Z}_\text{rate}^{l}$ are fed into multilayer perceptron (MLP) with softmax to get the $l$-th decision score $\mathbf{D}_{l}\in{\mathbb{R}^{l_{v}\times q}}$:

\begin{equation}
    \mathbf{D}_{l} = \operatorname{Softmax}(\operatorname{MLP}(\mathbf{Z}_\text{rate}^{l})),
\end{equation}
where $\operatorname{MLP}(\cdot)$ is MLP with stacked linear layers with $\operatorname{GeLU}(\cdot)$ activation function. $\operatorname{Softmax}(\cdot)$ denotes the softmax operation.

For the last rate predictor, all previous $\mathbf{D}_{l}$($l\in\left [1,L-1\right ]$) are used as additional inputs to help this rate predictor make the final decision $\mathbf{D}\in{\mathbb{R}^{l_{v}\times q}}$. The final prediction process can be formulated as:
\begin{equation}
   \mathbf{D} = \operatorname{Softmax}(\operatorname{MLP}(\operatorname{concat}(\mathbf{Z}_\text{rate}^{L},\beta(\mathbf{D}_{1}\dots\mathbf{D}_{L-1})))),
\end{equation}
where $\beta(\cdot)$ is the function that aggregates the previous decision. $\beta(\cdot)$ can be attention-based aggregation or other aggregation methods. For simplicity, We utilize the average operation to implement $\beta(\cdot)$:
\begin{equation}
    \beta(\mathbf{D}_{1}\dots\mathbf{D}_{L-1})=\frac{1}{L}{\sum_{i=1}^{L-1}}\mathbf{D}_{i}.
\end{equation}

Here, $\mathbf{D}$ represents the probability that each $\mathbf{Y}_{v,i}$ is classified into $q$ different fixed bandwidths. Then, we need to introduce the Gumbel-Softmax trick to solve the non-differentiable sampling problem, which is often used in network pruning. 

Given the $d$-dimensional token $\mathbf{Y}_{v,i}$ ($i\in\left [ 1,l_{v} \right ]$), we want to draw the sample $\mathbf{P}_{i}^\text{hard}\in{\mathbb{R}^{q}}$ representing the chosen bandwidth from a categorical distribution with the class probability ${\mathbf{D}_{i}}\in{\mathbb{R}^{q}}$. First, the Gumbel-Max trick formulates the sampling process as:
\begin{equation}
    \mathbf{P}_{i}^\text{hard} = \operatorname{onehot}(\mathop{\arg\max}\limits_{j}({\log(\mathbf{D}_{i,j})+\mathbf{G}_{i,j}})),\;j\in\left [ 1,q \right ], 
\end{equation}
where all elements of $\mathbf{G}\in{\mathbb{R}^{l_{v}\times{q}}}$ follow the Gumbel distribution $\operatorname{Gumbel}(0,1)$ and $\operatorname{onehot}(\cdot)$ is the one-hot encoding function. $\mathbf{G}_{i,j}$ can be computed by:
\begin{equation}
\mathbf{G}_{i,j}=-\log(-\log(\mathbf{U}_{i,j})),  
\end{equation}
where $\mathbf{U}\in{\mathbb{R}^{l_{v}\times{q}}}$ consists i.i.d. samples drawn from $\operatorname{Uniform}(0,1)$. Then, the softmax function with temperature coefficient $\tau$ is used as a continuous, differentiable approximation to $\arg \max_{\ }(\cdot)$, obtaining the soft version $\mathbf{P}_{i}^\text{soft}\in{\mathbb{R}^{q}}$ of $\mathbf{P}_{i}^\text{hard}$:
\begin{equation}
    \mathbf{P}_{i}^\text{soft}= \frac{e^{(\log(\mathbf{D}_{i,j})+\mathbf{G}_{i,j})/\tau} }{{\sum_{j=1}^{q}}e^{(\log(\mathbf{D}_{i,j})+\mathbf{G}_{i,j})/\tau}}.
\end{equation}

Through the Gumbel-Softmax trick, we involve $\mathbf{P}_{i}^\text{hard}$ in the forward propagation of the network, however, during backpropagation, we update the parameters by computing the gradient of $\mathbf{P}_{i}^\text{soft}$. As the temperature coefficient $\tau$ decreases, the soft version $\mathbf{P}_{i}^\text{soft}$ becomes closer to the hard version $\mathbf{P}_{i}^\text{hard}$, which gradually aligns the forward and backward propagation processes of the network. However, small $\tau$ can lead to instability of training. Therefore, we choose a large temperature coefficient $\tau$ at the beginning and gradually decay it during the training process. We use $\mathbf{P}_{i}^\text{hard}$ to select the corresponding bandwidth for each token $\mathbf{Y}_{v,i}$, indicating the number of retained channels. Note that for every $\mathbf{Y}_{v}\in{\mathbb{R}^{l_{v}\times d}}$, a corresponding $\mathbf{b}\in{\mathbb{R}^{l_{v}}}$ need to be transmitted through the lossless link to indicate the number of retained channels for each token at the receiver. $\mathbf{S}_{v}$ is generated by masking part of channels in $\mathbf{Y}_{v}$ as zero according to $\mathbf{M}$.

At the receiver, we first generate $\mathbf{M}$ based on received $\mathbf{b}$. Then, we unflatten and zero-pad noisy real channel output symbols $\hat{\mathbf{s}}_{v}$ based on $\mathbf{M}$ to get $\hat{\mathbf{S}}_{v}$. The learnable vector $\mathbf{c}\in\mathbb{R}^{d}$ is developed to compensate for the information loss due to the channel masking operation. For each token $\hat{\mathbf{S}}_{v,i}\in\mathbb{R}^{d}$, if its $j$-th channel is masked, $\mathbf{c}_{j}$ is selected as the initial value for this channel:
\begin{equation}
    \hat{\mathbf{S}}_{v,i}+(\mathbf{J}_{i}-\mathbf{M}_{i})\odot\mathbf{c}\to\hat{\mathbf{S}}_{v,i},
\end{equation}
where $\mathbf{J}\in\mathbb{R}^{l_{v}\times d}$ denotes the matrix whose all elements are set to $1$. Then, $\hat{\mathbf{S}}_{v}$ and $\mathbf{Y}_\text{rate}$ are inputted to the JSC decoder $f_{d}(\cdot;\boldsymbol{\phi})$ to decode the video semantics $\hat{\mathbf{Y}}_{v}$ progressively.

\subsection{Content-Adaptive and SNR-Adaptive}
The bandwidth allocation method described in \Cref{Section:bandwidth allocation} can achieve bandwidth efficiency for VideoQA-SC. The decision $\mathbf{D}$ is only determined by a series versions of $\mathbf{Y}_\text{rate}$ ($\mathbf{Y}_\text{rate}^{1}$,$\mathbf{Y}_\text{rate}^{2}$,...,$\mathbf{Y}_\text{rate}^{L}$) and a series versions of $\mathbf{Y}_v$ ($\mathbf{Y}_v^{1}$,$\mathbf{Y}_v^{2}$,...,$\mathbf{Y}_v^{L-1}$), which indicates that the bandwidth allocation is adaptive to video semantics or video contents. Such rate predictors will make the same bandwidth allocation under different channel conditions, which is inconsistent with traditional channel coding ideas. Since content-adaptive bandwidth allocation is not robust to noise, it is difficult to support SC under diverse channel conditions.

Assume that the transmitter can obtain the perfect SNR via ideal channel estimation. By introducing channel SNR, the rate predictors can integrate video content with the current channel condition for more reasonable bandwidth allocation not only adaptive to video contents but also to SNR. Specifically, we take SNR as the additional input of all rate predictors to make decisions adapt to the current channel condition at each layer of the JSC encoder.
    
For SNR-adaptive bandwidth allocation, $\mathbf{D}_{l}$ is computed by:
\begin{equation}
    \mathbf{D}_{l} = \operatorname{MLP}(\mathbf{Z}_\text{rate}^{l}, \text{SNR}),
\label{eq_17_new}
\end{equation}
and $\mathbf{D}$ is computed by:
\begin{equation}
   \mathbf{D} = \operatorname{MLP}(\operatorname{concat}(\mathbf{Z}_\text{rate}^{L},\beta(\mathbf{D}_{1}\dots\mathbf{D}_{L-1}),\text{SNR})).
\label{eq_18_new}
\end{equation}
In this way, as SNR is repeatedly used in rate predictors, the network is forced to learn dynamic bandwidth allocation strategies based on channel conditions, which enables VideoQA-SC robust to noise.
\subsection{Multimodal Fuser}
The multimodal fuser $u(\cdot;\boldsymbol{\nu})$ is used to interact the video and text information, and find the informative video contents with respect to the question for answer prediction.

Consider the process of a particular question-answer (QA) pair, such as the question $q(\textit{``what does the butterfly do 10 or more than 10 times''})$ and the candidate answers $a_{0}(\textit{``stuff marshmallow''})$, $a_{1}(\textit{``fall over''})$ and $a_{2}(\textit{``flap wings''})$. The text information can be organized as a 
tuple including three sequences $(q\left [ \operatorname{SEP} \right ] a_{0},q\left [ \operatorname{SEP} \right ] a_{1},q\left [ \operatorname{SEP} \right ] a_{2})$, which means each candidate answer is paired with the question to form a language sequence. $\left [ \operatorname{SEP} \right ]$ is a special sign used to separate the text of the question and the answer. Then, each sequence is transformed into tokens by the tokenizer. We use a language model to capture the correlations between each token in one sequence and extract the candidate QA-pair feature $\mathbf{Y}_{q}\in\mathbb{R}^{b\times{s_{i}}\times d}$ from $\mathbf{X}_{q}$, where $b$ is the number of candidate answers and $s_{i}$ is the length of $i$-th sequence.

Given $\hat{\mathbf{Y}}_{v}$ and $\mathbf{Y}_{q}$, the interaction of the two modal information can be achieved through attention-based weighted fusion. After linear projection, $\hat{\mathbf{Y}}_{v}$ is mapped to $\mathbf{E}_{v}\in\mathbb{R}^{l_{v}\times d}$ as query and $\mathbf{Y}_{q}$ is mapped to $\mathbf{E}_{q}\in\mathbb{R}^{b\times{s_{i}}\times d}$ as key. Then, we calculate the attention of $\hat{\mathbf{Y}}_{v}$ to each candidate QA pair feature $\mathbf{Y}_{q,i}$, and utilize attention-based fusion of $\hat{\mathbf{Y}}_{v}$ and $\mathbf{Y}_{q}$ to get the QA-aware video feature $\mathbf{Y}_{qv}\in\mathbb{R}^{l_{v}\times d}$, which can be formulated as:
\begin{align}
\begin{aligned}
    &\boldsymbol{\gamma}_{i} = \operatorname{softmax}(\mathbf{E}_{v}(\mathbf{E}_{q,i})^{T}), \\
    &\mathbf{Y}_{qv} = \hat{\mathbf{Y}}_{v}+ {\sum_{i=1}^{b}}\boldsymbol{\gamma}_{i}\mathbf{Y}_{q,i}.
\end{aligned}
\end{align}
\begin{algorithm}[!t]
\caption{Forward process of VideoQA-SC over the AWGN channel.}\label{alg:alg1}
\begin{algorithmic}[1]
 \Require The chosen mini-batch data $\left \{ (\mathbf{X}_{v},\mathbf{X}_{q}) \right \}_{i}^{i+bz}$; training channel SNR $\text{SNR}_{\text{train},i}$ and model parameters $\boldsymbol{\zeta}$, $\boldsymbol{\theta}$, 
 $\boldsymbol{\epsilon}$, $\boldsymbol{\phi}$, $\boldsymbol{\nu}$.
 \Ensure  the answer score $\left \{\mathbf{a}_\text{pred} \right \}_{i}^{i+bz}$ and the binary mask matrix $\mathbf{M}$.
  \State $g(\left \{ \mathbf{X}_{v} \right \}_{i}^{i+bz};\boldsymbol{\zeta})\to \left \{ \mathbf{Y}_{v} \right \}_{i}^{i+bz}$.
  \If{$\text{SNR}_{\text{train},i}=$None}
  \State $f_{e}(\left \{ \mathbf{Y}_{v} \right \}_{i}^{i+bz};\boldsymbol{\theta})\to \left \{ \mathbf{S}_{v} \right \}_{i}^{i+bz}$ and generate $\left \{\mathbf{M} \right \}_{i}^{i+bz}$.
  \Else 
  \State $f_{e}(\left \{ \mathbf{Y}_{v} \right \}_{i}^{i+bz},\text{SNR}_\text{train};\boldsymbol{\theta},\boldsymbol{\epsilon})\to \left \{ \mathbf{S}_{v} \right \}_{i}^{i+bz}$ and generate $\left \{\mathbf{M} \right \}_{i}^{i+bz}$.
 \EndIf
 \State Flatten and real-to-complex: $\left \{ \mathbf{S}_{v} \right \}_{i}^{i+bz} \to \left \{\mathbf{s}_{v}^{c} \right \}_{i}^{i+bz}$ .
 \State Compute the average noise power $\sigma^{2}$ by $\text{SNR}_{\text{train},i}$ and sample $\mathbf{n}$ from $\mathcal{CN}(0,\sigma^{2}\mathbf{I})$ $bz$ times.
 \State $\left \{ \hat{\mathbf{s}}_{v}^{c} \right \}_{i}^{i+bz}$ = $\left \{ \mathbf{s}_{v}^{c} \right \}_{i}^{i+bz} + \left \{ \mathbf{n} \right \}_{i}^{i+bz}$.
 \State Complex-to-real and unflatten: $\left \{ \hat{\mathbf{s}}_{v}^{c} \right \}_{i}^{i+bz} \to \left \{ \hat{\mathbf{S}}_{v} \right \}_{i}^{i+bz}$.
 \State $f_{d}\left(\left \{ \hat{\mathbf{S}}_{v} \right \}_{i}^{i+bz};\boldsymbol{\phi}\right)\to \left \{ \hat{\mathbf{Y}}_{v} \right \}_{i}^{i+bz}$.
 \State $u\left(\left \{ \hat{\mathbf{Y}}_{v} \right \}_{i}^{i+bz},\left \{ \mathbf{X}_{q} \right \}_{i}^{i+bz};\boldsymbol{\nu}\right) \to \left \{\mathbf{a}_\text{pred} \right \}_{i}^{i+bz}$.
 \State\Return $\left \{\mathbf{a}_\text{pred} \right \}_{i}^{i+bz}$, $\left \{\mathbf{M} \right \}_{i}^{i+bz}$.
\end{algorithmic}
\end{algorithm}
\begin{algorithm}[t]
 \caption{Training of the SNR-adaptive VideoQA-SC.}\label{alg:alg2}
 \begin{algorithmic}[1]
 \Require the training dataset $\mathcal{D}_\text{train}$; batchsize $bz$; trained VideoQA model parameters $\boldsymbol{\zeta}$ and $\boldsymbol{\nu}$; the trade-off hyperparameter $\lambda$.
 \Ensure  the trained parameters $\boldsymbol{\zeta}^{\star}$, $\boldsymbol{\theta}^{\star}$,
 $\boldsymbol{\epsilon}^{\star}$,
 $\boldsymbol{\phi}^{\star}$,
 $\boldsymbol{\nu}^{\star}$.
  \\ \textbf{Initialization} : DJSCC transmission model parameters $\boldsymbol{\theta}$, $\boldsymbol{\epsilon}$, $\boldsymbol{\phi}$; training SNR range: $(\text{SNR}_\text{start}, \text{SNR}_\text{end})$.

 \State Freeze $\boldsymbol{\zeta}$, $\boldsymbol{\nu}$ and  $\boldsymbol{\epsilon}$. \Comment{\textbf{fixed-bandwidth DJSCC training}}

 \State Choose mini-batch data $\left \{ (\mathbf{X}_{v},\mathbf{X}_{q},\mathbf{a}_\text{label}) \right \}_{i}^{i+bz}$.
 \State Sample $\text{SNR}_{\text{train},i}$ from $\operatorname{Uniform}(\text{SNR}_\text{start},\text{SNR}_\text{end})$.
 \State Compute $\left \{\mathbf{a}_\text{pred} \right \}_{i}^{i+bz}$ by Algorithm 1 \textit{without input} $\text{SNR}_{\text{train},i}$.
 \State Compute $\mathcal{L}_\text{stage2}$ by Eq. (\ref{eq_stage2}) and optimize $\boldsymbol{\theta}$ and $\boldsymbol{\phi}$ by gradient descent.
 \State Repeat line 3-6 to complete the training of stage 2.
 \State Unfreeze $\boldsymbol{\epsilon}$. \Comment{\textbf{bandwidth-adaptive DJSCC training}}
 \State Repeat line 3-4.
 \State Compute $\left \{\mathbf{a}_\text{pred} \right \}_{i}^{i+bz}$ and $\left \{\mathbf{M} \right \}_{i}^{i+bz}$ by Algorithm 1 \textit{with input} $\text{SNR}_{\text{train},i}$.
  \State Compute $\mathcal{L}_\text{stage3}$ by Eq. (\ref{eq_stage3}) and optimize $\boldsymbol{\theta}$, $\boldsymbol{\phi}$ and $\boldsymbol{\epsilon}$ by gradient descent.
 \State Repeat line 9-11 to complete the training of stage 3.
 \State Unfreeze $\boldsymbol{\zeta}$ and $\boldsymbol{\nu}$. \Comment{\textbf{VideoQA-SC finetuning}}
 \State Repeat line 9-10.
 \State Compute $\mathcal{L}_\text{stage4}$ by Eq. (\ref{eq_stage4}) and optimize all parameters by gradient descent.
 \State Repeat line 14-15 to complete the training of stage 4.
 \State\Return Optimized $\boldsymbol{\zeta}^{\star}$, $\boldsymbol{\theta}^{\star}$,
 $\boldsymbol{\epsilon}^{\star}$,
 $\boldsymbol{\phi}^{\star}$,
 $\boldsymbol{\nu}^{\star}$.
 \end{algorithmic}
 \end{algorithm}

We add some Transformer block to refine the QA-aware feature $\mathbf{Y}_{qv}$ and average pool it along the token dimension to get the global QA-aware video feature $\mathbf{Y}_{qv}^\text{global}\in\mathbb{R}^{1\times d}$. Similarly, the candidate QA-pair feature $\mathbf{Y}_{q}$ is average pooled along the sequence length dimension to get the global text feature $\mathbf{Y}_{q}^\text{global}\in\mathbb{R}^{b \times d}$.
We simply utilize the dot product with softmax to obtain the answer score $\mathbf{a}_\text{pred}\in\mathbb{R}^{1\times b}$ by measuring the similarity between $\mathbf{Y}_{q}^\text{global}$ and $\mathbf{Y}_{qv}^\text{global}$:
\begin{equation}
    \mathbf{a}_\text{pred} = \operatorname{softmax}(\mathbf{Y}_{qv}^\text{global}(\mathbf{Y}_{q}^\text{global})^{T}).
\end{equation}
Finally, the answer with the highest prediction score is output as the predicted answer $a^{\star}$ by the multimodal fuser $u(\cdot;\boldsymbol{\nu})$:
\begin{equation}
    a^\star = \mathop{\arg\max}\limits_{i}{\mathbf{a}}_{\text{pred},i},
\end{equation}
where ${\mathbf{a}}_{\text{pred},i}$ is the $i$-th element of $\mathbf{a}_\text{pred}$.

\subsection{Training Strategy}
VideoQA-SC can be considered as a combination of the VideoQA model ($g(\cdot;\boldsymbol{\zeta})$, $u(\cdot;\boldsymbol{\nu})$) and the DJSCC transmission model ($f_{e}(\cdot;\boldsymbol{\theta}, \boldsymbol{\epsilon})$, $f_{d}(\cdot;\boldsymbol{\phi})$), where the VideoQA model is responsible for accurate execution of VideoQA tasks and the DJSCC model is responsible for reliable and efficient transmission of video semantics. Instead of E2E training the whole system from scratch, we adopt a progressive training strategy to ensure the stability of training process. The progressive training strategy mainly includes 4 stages:
\subsubsection{VideoQA Model Training} Train the VideoQA model $g(\cdot;\boldsymbol{\zeta})$ and $u(\cdot;\boldsymbol{\nu})$ without considering the noisy channel. The video semantics extracted by the video semantic encoder $g(\cdot;\boldsymbol{\zeta})$ are losslessly inputted into multimodal fuser $u(\cdot;\boldsymbol{\nu})$ to predict the answer. The cross-entropy between the prediction of the VideoQA model and the one-hot form label is used as the task loss $\mathcal{L}_\text{task}$ in this stage:
\begin{equation}
\mathcal{L}_\text{stage1}=\mathcal{L}_\text{task}=\operatorname{CE}(\mathbf{a}_\text{pred},\mathbf{a}_\text{label}),
\label{eq_stage1}
\end{equation}
where $\operatorname{CE}(\cdot,\cdot)$ is the cross-entropy loss.

\subsubsection{Fixed-Bandwidth DJSCC Transmission Training} Train the fixed-bandwidth DJSCC transmission model ($f_{e}(\cdot;\boldsymbol{\theta}, \boldsymbol{\epsilon})$ and $f_{d}(\cdot;\boldsymbol{\phi})$) without rate predictors parameters $\boldsymbol{\epsilon}$ while freezing VideoQA model parameters $\boldsymbol{\zeta}$ and $\boldsymbol{\nu}$. Taking the noisy channel into account, the training objective of the fixed-bandwidth DJSCC transmission model is to maximize the task performance. This stage has the same loss as $\mathcal{L}_\text{stage1}$.
\begin{equation}
\mathcal{L}_\text{stage2}=\mathcal{L}_\text{task}=\operatorname{CE}(\mathbf{a}_\text{pred},\mathbf{a}_\text{label}).
\label{eq_stage2}
\end{equation}

\subsubsection{Bandwidth-Adaptive DJSCC Transmission Training}
Train the bandwidth-adaptive DJSCC transmission model ($f_{e}(\cdot;\boldsymbol{\theta}, \boldsymbol{\epsilon})$ and $f_{d}(\cdot;\boldsymbol{\phi})$) with parameters of rate predictors $\boldsymbol{\epsilon}$. The training objective of this stage can be formulated as the trade-off between the task performance and the bandwidth cost. Therefore, the bandwidth cost loss $\mathcal{L}_\text{rate}$ with the hyperparameter $\lambda$  is introduced in this stage:
\begin{equation}
\begin{aligned}
\mathcal{L}_\text{stage3}&=\mathcal{L}_\text{task}+\lambda\mathcal{L}_\text{rate}\\&=\operatorname{CE}(\mathbf{a}_\text{pred},\mathbf{a}_\text{label})+\lambda\sum_{i}\sum_{j}\mathbf{M}_{i,j},
\end{aligned}
\label{eq_stage3}
\end{equation}
where $\mathbf{M}_{i,j}$ denotes the $i$-th row and $j$-th column element in the binary mask matrix $\mathbf{M}$, and $\lambda$ controls the trade-off between the task performance and the bandwidth cost.
\subsubsection{E2E Finetuning}Unfreeze VideoQA model parameters $\boldsymbol{\zeta}$ and $\boldsymbol{\nu}$. Jointly finetune all system parameters $\boldsymbol{\zeta}$, $\boldsymbol{\theta}$, $\boldsymbol{\epsilon}$, $\boldsymbol{\phi}$ and $\boldsymbol{\nu}$ to improve the E2E system performance. This stage has the same loss as $\mathcal{L}_\text{stage3}$:
\begin{equation}
\begin{aligned}
\mathcal{L}_\text{stage4}&=\mathcal{L}_\text{task}+\lambda\mathcal{L}_\text{rate}\\&=\operatorname{CE}(\mathbf{a}_\text{pred},\mathbf{a}_\text{label})+\lambda\sum_{i}\sum_{j}\mathbf{M}_{i,j}.
\end{aligned}
\label{eq_stage4}
\end{equation}

\Cref{alg:alg1} demonstrates the forward process of VideoQA-SC over the AWGN channel. Furthermore, taking SNR information into bandwidth allocation, the training process for SNR-adaptive VideoQA-SC is shown in \Cref{alg:alg2}.

\section{Experiments}\label{Section:Experiments}
In this section, we introduce the experimental setup and provide quantified experimental results to demonstrate the effectiveness of VideoQA-SC for performing VideoQA tasks. We compare VideoQA-SC with traditional systems based on separate source channel coding (SSCC) and SC systems adopted advanced DJSCC transmission schemes under various channel conditions and bandwidth constraints.
\subsection{Experimental Setup}
\subsubsection{Datasets} We choose the TGIF-QA dataset as the benchmark to conduct our experiments. As one of the popular datasets for VideoQA, TGIF-QA dataset consists of $165165$ QA pairs chosen from $71741$ animated GIFs. To evaluate the spatiotemporal reasoning ability at the video level, TGIF-QA dataset designs four unique task types, i.e., repetition count, repeating action, state transition and frame QA. 

We select repeating action and state transition for experiments, which are the two most challenging tasks in the TGIF-QA dataset. The two tasks are defined as multiple choice questions. Each question has $5$ candidate answers. The questions for repeating action involves identifying the repeated action in a video. The questions for state transition involves identifying the state before or after a particular state, including facial expressions, actions, places and object properties. For convenience, we refer to the repeating action task and state transition in the TGIF-QA dataset as the TGIF-QA Action dataset and TGIF-QA Transition dataset, respectively.

To overcome serious language bias in the original TGIF-QA dataset, we use questions and answers from an enhanced version of TGIF-QA, i.e., TGIF-QA-R \cite{TGIF-QA-R} to force reasoning based on both text and video content. TGIF-QA-R has $20475$ QA pairs and $2274$ QA pairs as training and testing datasets for repeating action task, respectively. It has $52704$ QA pairs and $6232$ QA pairs as training and testing datasets for state transition task, respectively.

\subsubsection{Implementation Details}
In our experiments, $l_{v}$, ${l_{c}}$, ${l_{f}}$ are set to $16$, $4$, $4$ for sparse sampling. $r$ and $m$ are set to $10$ and $2048$ for spatiotemporal semantic encoder. Note that the pretrained object-level feature extractor and the pretrained frame-level feature extractor served as preprocessing networks are not involved in the training process. $r$ is set to $10$ means that we only select the top $10$ detected objects with the highest confidence for each frame, which is a trade-off between network complexity and network performance. When the number of detected objects is less than $10$, we pad the additional object features with zeros to ensure a total of $10$ detected objects. $d$ is set to $256$ for DJSCC transmission, which also represents the maximum bandwidth that can be allocated for each token. Both the JSC encoder and the JSC decoder consist of $L=4$ Transformer blocks for JSC encoding/decoding. There are also $L=4$ rate predictors for bandwidth allocation during JSC encoding. $q$ is set to $8$ for adaptive bandwidth allocation, which denotes the set of candidate retained channels for each token is $\mathcal{R}=\left \{2,4,8,16,32,64,128,256 \right \}$. Each question has $b=5$ candidate answers to form QA pairs.

In progressive training of VideoQA-SC, the training epochs for stage 1, stage 2, stage 3 and stage 4 are $20$, $20$, $20$ and $10$, respectively. The learning rates for stage 1, stage 2, stage 3 and stage 4 are $1\times{10}^{-5}$, $5\times{10}^{-6}$, $5\times{10}^{-6}$ and $2\times{10}^{-6}$, respectively. The Gumbel-Softmax trick is enabled in the training stage 3 and stage 4 for differentiable sampling. For stable training, we set the temperature coefficient $\tau=5$ at the beginning and decay it by a factor of $0.9$ after each epoch in stage 3. Similarly, we set $\tau=1$ initially and decay it by a factor of $0.95$ after each epoch in stage 4. We train models that satisfy different bandwidth constraints by tuning the hyperparameter $\lambda$.

\subsubsection{Comparison Schemes}
We compare the VideoQA-SC with SSCC-based traditional systems and DJSCC-based SC systems that perform VideoQA based on reconstructed videos. Specifically, we use SSCC and DJSCC schemes to transmit videos over the noisy wireless channel, and use the same optimized VideoQA model ($g(\cdot;\boldsymbol{\zeta})$ and $u(\cdot;\boldsymbol{\nu})$) to perform VideoQA tasks. 

For SSCC-based traditional systems, we adopt the traditional video codecs ($\text{H}264/265$) for source coding and assume that the channel capacity is achievable to obtain the upper bound of performance. The SSCC comparison schemes consist of ``$\text{H}264 + \text{channel capacity}$" and ``$\text{H}265 + \text{channel capacity}$", which denotes the combination of $\text{H}264$ or $\text{H}265$ and the optimal channel coding achieving channel capacity. FFmpeg is adpoted to simulate the video coding process of $\text{H}264$ and $\text{H}265$.

For DJSCC-based SC systems, we adopt the advanced DJSCC-based video transmission model DVST\cite{video2} to transmit videos. Since DVST only focuses on the coding of P-frames, DJSCC-based image compression neural networks are required to encode I-frames. For a simple and fair comparison, we assume that I-frames can be transmitted losslessly and DVST is used to encode P-frames based on lossless I-frames. We only consider the average bandwidth of P-frames. Then, the performance upper bound of the DJSCC-based SC system is obtained through the optimal video reconstruction with the minimum bandwidth.

\begin{figure*}[t]
\centering
\subfloat[The content-adaptive method in TGIF-QA-R Action dataset.]{
\centering
\includegraphics[width=0.453\linewidth]{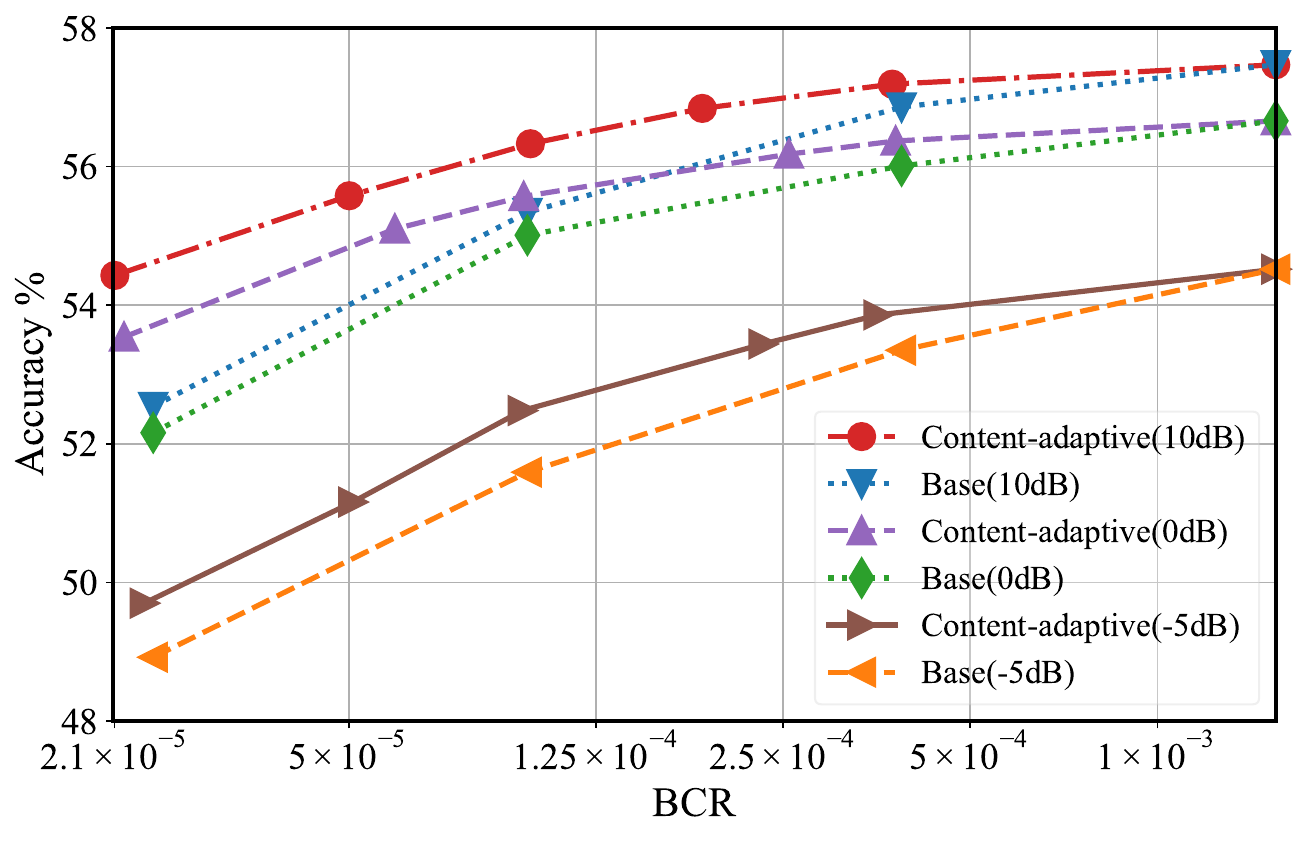}%
\label{Fig:A_cbr}}
\hfil
\subfloat[The content-adaptive method in TGIF-QA-R Transition dataset.]{
\centering
\includegraphics[width=0.46\linewidth]{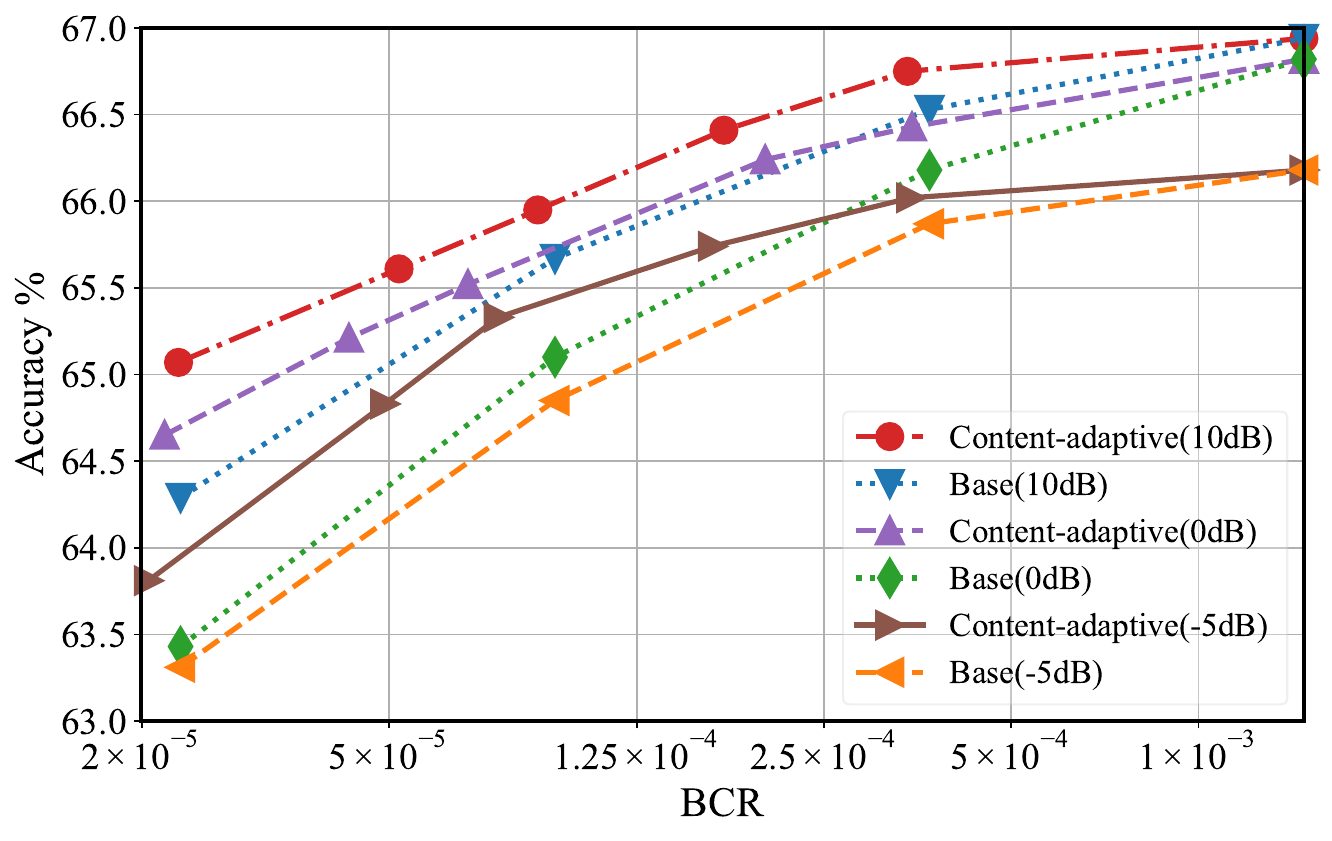}%
\label{Fig:T_cbr}}
\caption{Answer accuracy of different versions of VideoQA-SC versus the average BCR under fixed SNRs over the AWGN channel. Each line is trained with a particular SNR.}
\label{Fig:cbr}
\end{figure*}
\begin{figure*}[t]
\centering
\subfloat[The SNR-adaptive method in TGIF-QA-R Action dataset.]{\includegraphics[width=0.453\linewidth]{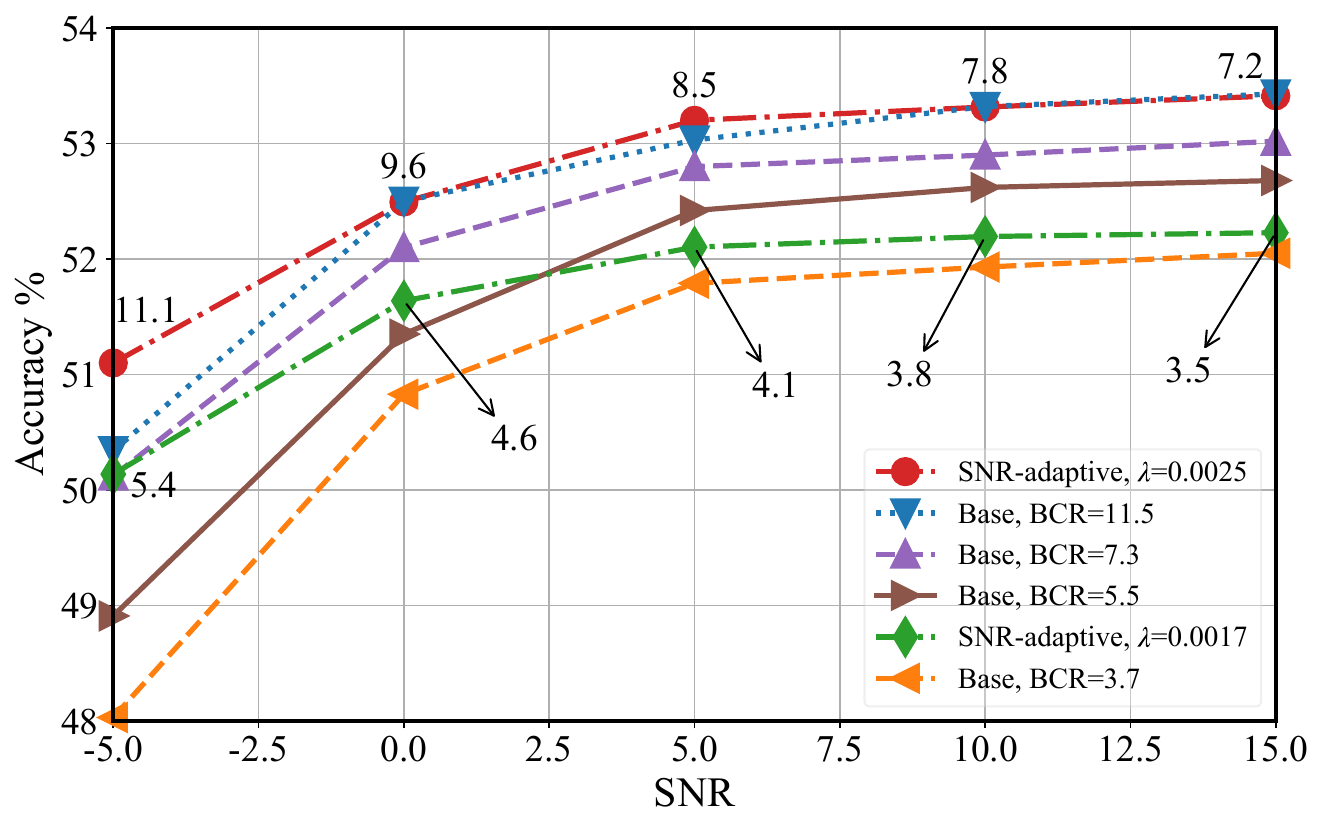}%
\label{Fig:A_snr}}
\hfil
\subfloat[The SNR-adaptive method in TGIF-QA-R Transition dataset.]{\includegraphics[width=0.46\linewidth]{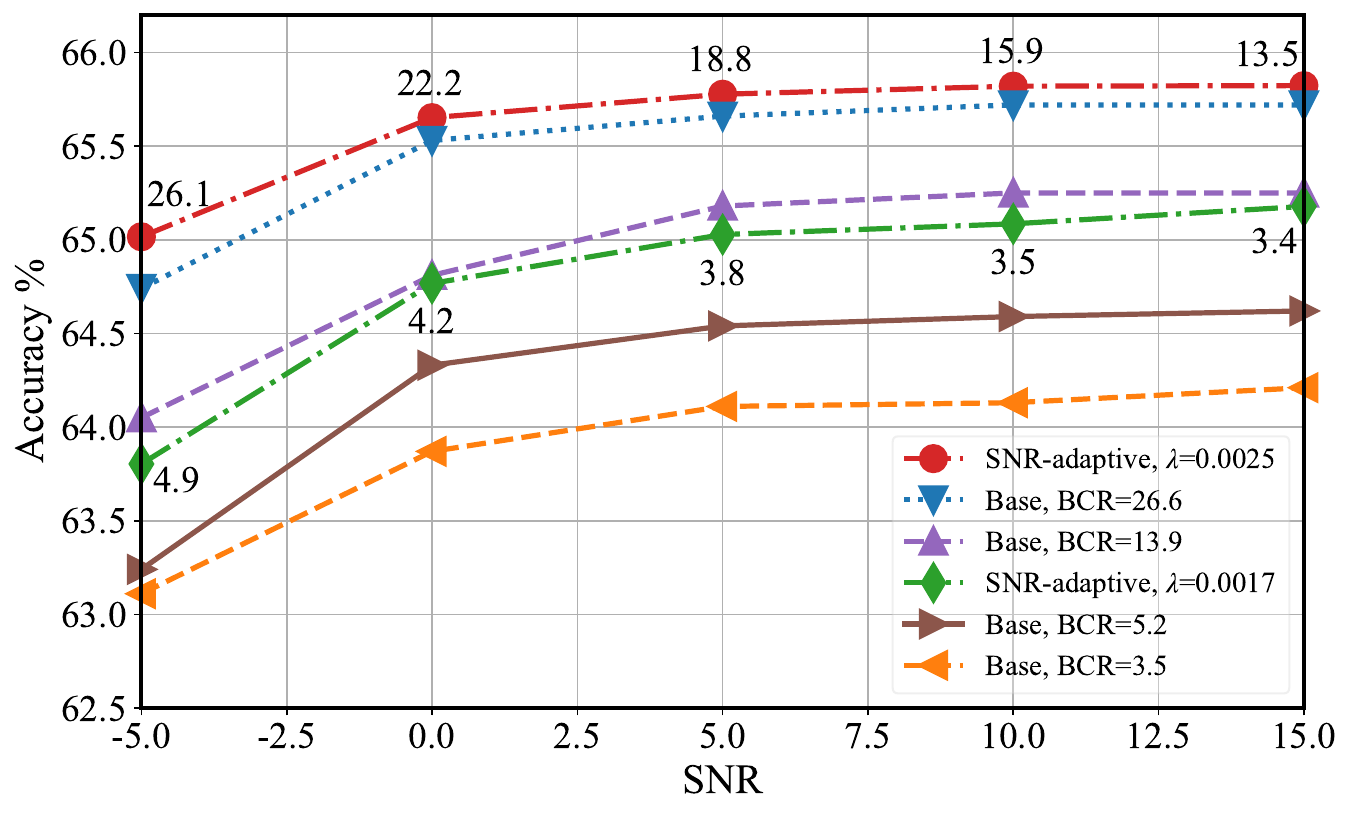}%
\label{Fig:T_snr}}
\caption{Answer accuracy of different versions of VideoQA-SC versus SNRs over the AWGN channel.}
\label{Fig:snr}
\end{figure*}

\subsection{Ablation Study}
VideoQA-SC enables dynamic bandwidth allocation under the guidance of multiple information to fully leverage limited bandwidth resources. We perform ablation experiments based on two bandwidth-adaptive methods (content-adaptive and SNR-adaptive bandwidth allocation) to verify their effectiveness. 

We train and test our models over AWGN channels under fixed SNRs and mixed SNRs to validate the effectiveness of the two bandwidth-adaptive methods, respectively. The mixed training SNR range is from $-5$ to $15$ dB. Fig. \ref{Fig:cbr}\subref{Fig:A_cbr} and \subref{Fig:T_cbr} show the accuracy of different versions of VideoQA-SC constrained by various BCRs under $3$ SNRs ($-5$, $0$, $10$ dB). For each particular SNR, a content-adaptive VideoQA-SC is compared with a base VideoQA-SC which does not utilize rate predictors for bandwidth allocation.

It can be observed that under different BCR constraints, the accuracy of content-adaptive VideoQA-SC is consistently higher than that of the base VideoQA-SC. As the BCR decreases, the accuracy gain of content-adaptive VideoQA-SC gradually increases. It indicates that when the available bandwidth is extremely constrained, the content-adaptive bandwidth allocation can effectively utilize limited bandwidth resources to improve the overall performance of VideoQA tasks. Moreover, content-adaptive bandwidth allocation does not consider channel conditions, which allocates bandwidth to different tokens only based on video semantics. In scenarios with high SNRs, e.g., $10$ dB, the accuracy gains from content-adaptive bandwidth allocation become more pronounced because video semantics can be transmitted more accurately with less channel noise. In particular, when the SNR is $10$ dB and the BCR is constrained to $2.4\times 10^{-5}$ (the average retained channels for each token is $4$), the accuracy gain is about $1.89\%$ for the content-adaptive VideoQA-SC in the TGIF-QA-R Action dataset.

By incorporating the estimated SNR into the rate predictors, VideoQA-SC can jointly consider video semantics and current channel conditions to make decisions, which is called SNR-adaptive bandwidth allocation. Fig. \ref{Fig:snr}\subref{Fig:A_snr} and \subref{Fig:T_snr} illustrate the different bandwidth allocations of SNR-adaptive VideoQA-SC under various SNRs. For readability, The average BCR for each point in lines representing SNR-adaptive VideoQA-SCs is normalized by $10^{-5}$ and marked near this point.

By observing lines representing SNR-adaptive VideoQA-SCs, a basic trend can be seen that as the SNR decreases, the SNR-adaptive VideoQA-SC tends to allocate more bandwidth for transmitting videos to resist stronger channel noise. Furthermore, two fixed-bandwidth base VideoQA-SCs are used for comparison with SNR-adaptive VideoQA-SCs with different trade-off parameter $\lambda$. We train the two fixed-bandwidth VideoQA-SCs based on the highest and lowest allocated bandwidths of the SNR-adaptive VideoQA-SCs under different SNRs. As SNR decreases from $5$ dB to $-5$ dB, the base VideoQA-SCs experience rapid performance degradation due to the fixed bandwidth allocation. In contrast, the SNR-adaptive VideoQA-SC is able to adaptively adjust the bandwidth allocation policy according to the current SNR, thereby achieving smooth performance degradation. Particularly, when the SNR is $-5$ dB, the SNR-adaptive VideoQA-SC achieves an accuracy gain up to $2.10\%$ compared with the base VideoQA-SC in the TGIF-QA-R Action dataset.
\begin{figure*}[t]
\centering
\subfloat[SNR$=0$ dB.]{\includegraphics[width=0.33\linewidth]{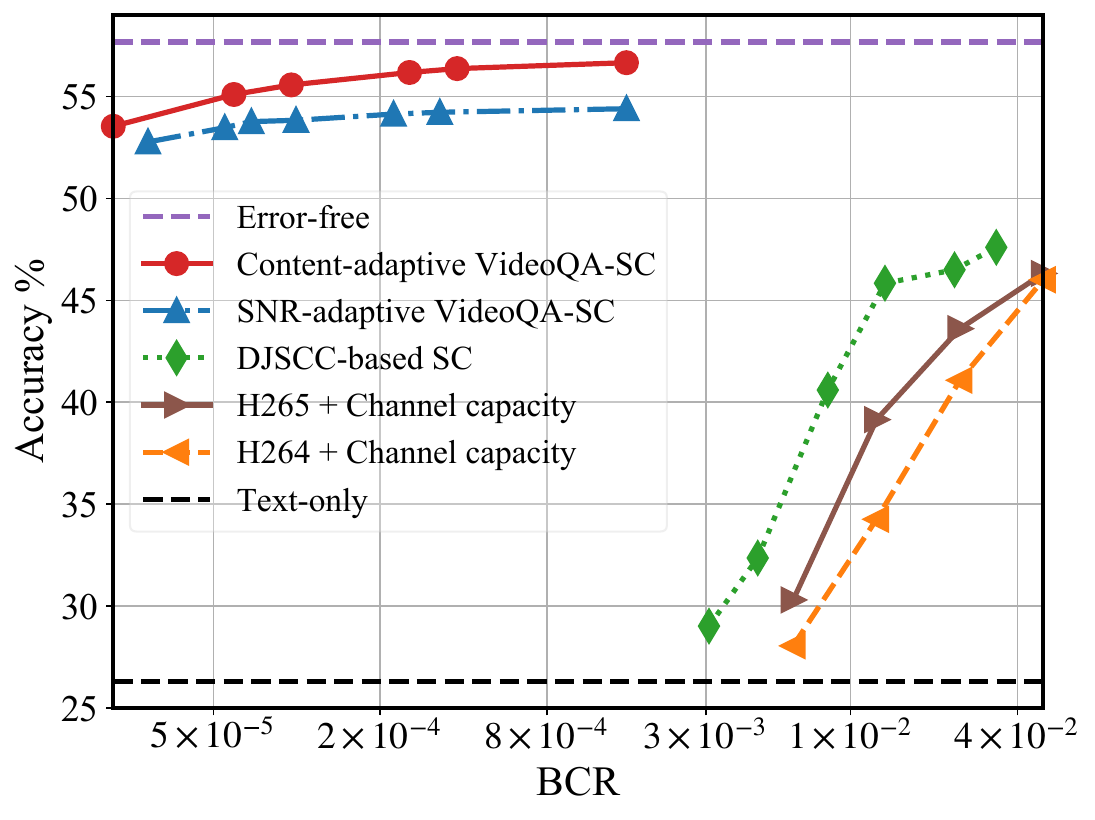}%
\label{Fig:A_0dB}}
\hfil
\subfloat[SNR$=5$ dB.]{\includegraphics[width=0.33\linewidth]{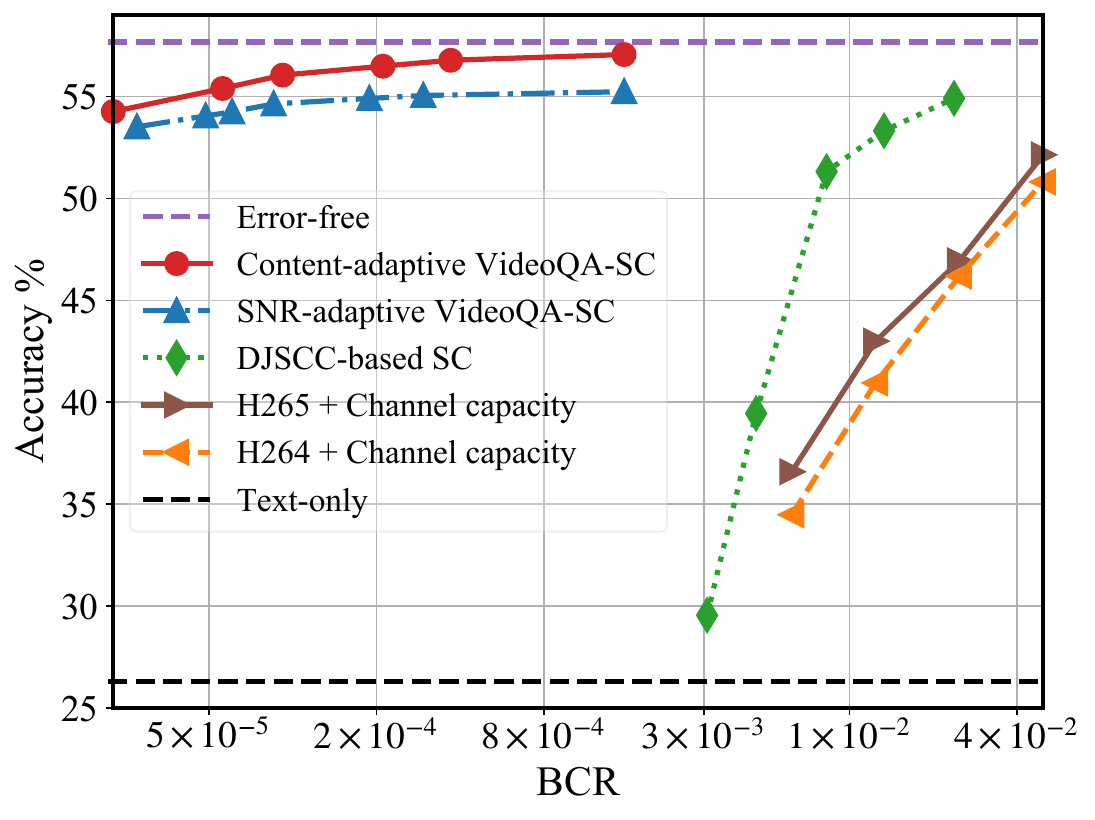}%
\label{Fig:A_5dB}}
\hfill
\subfloat[SNR$=10$ dB.]{\includegraphics[width=0.33\linewidth]{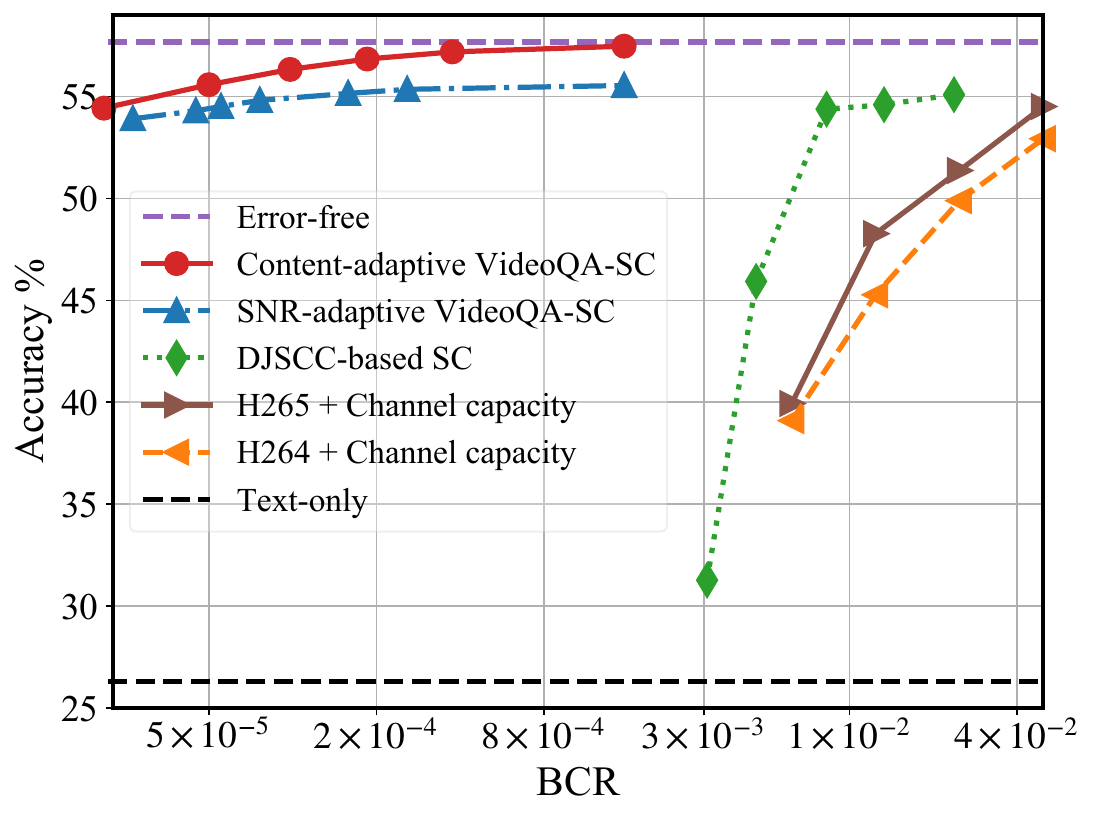}%
\label{Fig:A_10dB}}
\caption{Answer accuracy of different SC models versus average BCR over AWGN channels under different SNRs in TGIF-QA-R Action dataset.}
\label{Fig:Action_All}
\end{figure*}
\begin{figure*}[t]
\centering
\subfloat[SNR$=0$ dB.]{\includegraphics[width=0.33\linewidth]{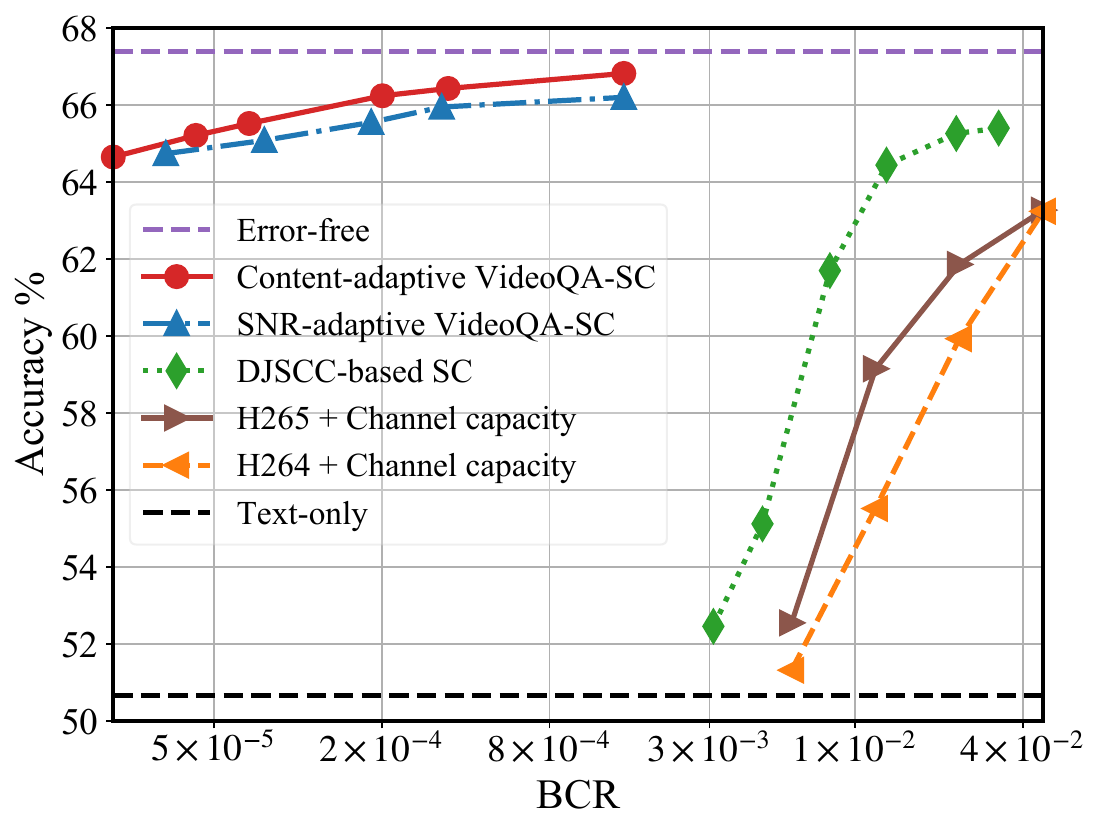}%
\label{Fig:T_0dB}}
\hfil
\subfloat[SNR$=5$ dB.]{\includegraphics[width=0.33\linewidth]{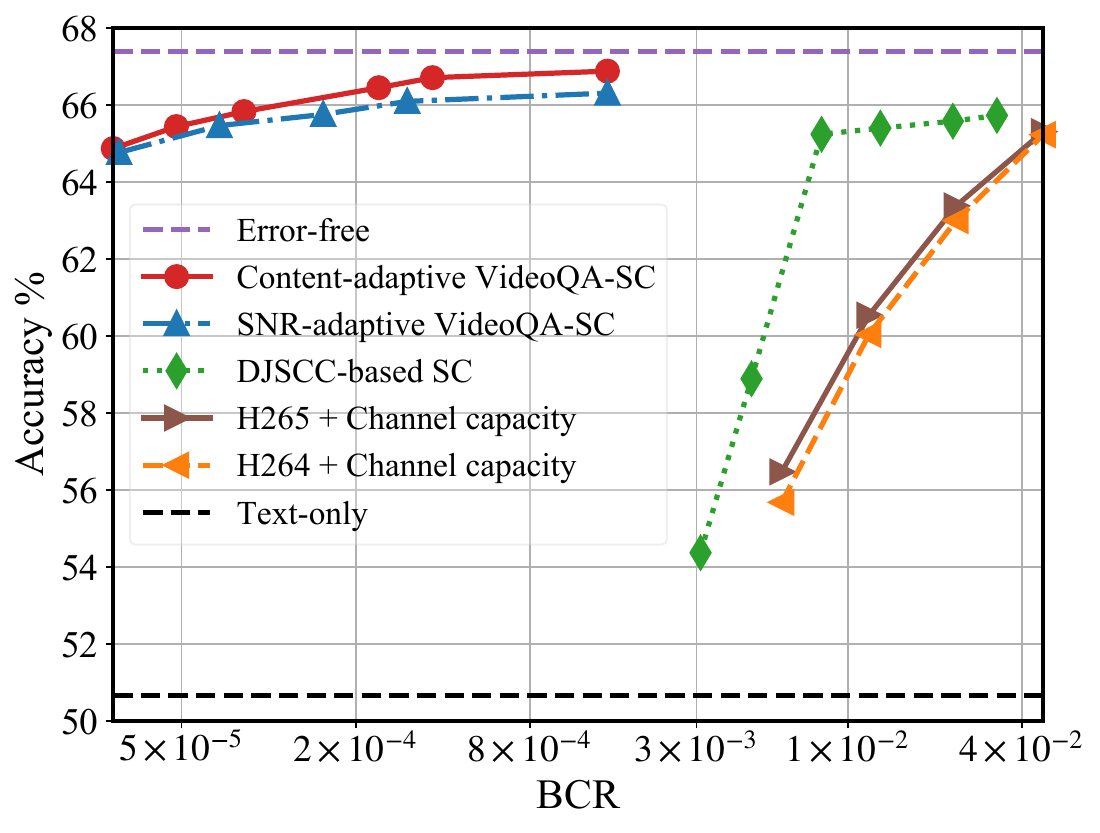}%
\label{Fig:T_5dB}}
\hfill
\subfloat[SNR$=10$ dB.]{\includegraphics[width=0.33\linewidth]{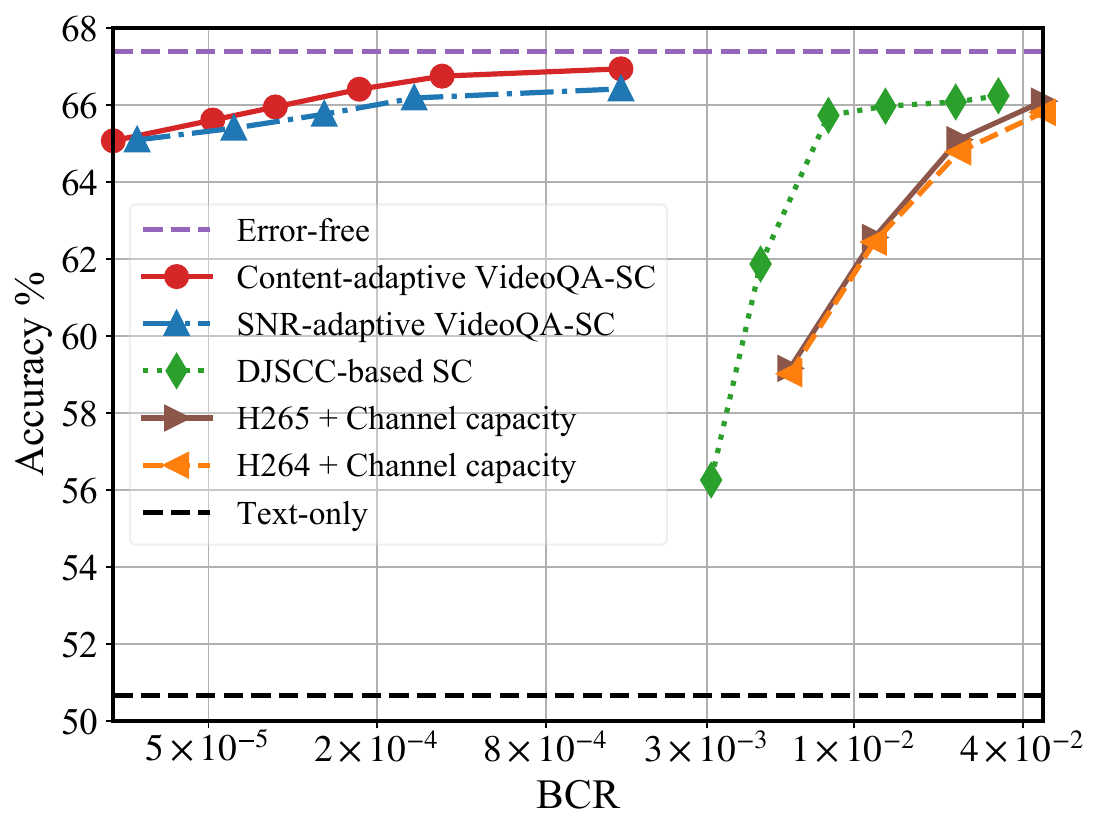}%
\label{Fig:T_10dB}}
\caption{Answer accuracy of different SC models versus average BCR over AWGN channels under different SNRs in TGIF-QA-R Transition dataset}
\label{Fig:Transition_All}
\end{figure*}

Another observation is that the SNR-adaptive VideoQA-SC not only dynamically adjusts the bandwidth allocation according to the SNR, but also outperforms the base VideoQA-SC with the same bandwidth in most cases. Fig. \ref{Fig:snr}\subref{Fig:T_snr} shows that SNR-adaptive VideoQA-SC outperforms even the base Video-SC with the highest fixed bandwidth under all SNRs in the TGIF-QA-R Transition dataset. This suggests that SNR-adaptive VideoQA-SC is still capable of accomplishing content-adaptive bandwidth allocation based on video semantics, which shows the powerful scalability of the proposed learning-based bandwidth allocation. By integrating SNR with video semantics for bandwidth allocation, SNR-adaptive VideoQA-SC achieves promising task performance while avoiding the need to train multiple models for specific channel conditions and the frequent model switching during practical application.

\subsection{Performance Comparison of Different Models}
As depicted in Fig. \ref{Fig:Action_All} and \ref{Fig:Transition_All}, the proposed VideoQA-SC is compared with SSCC-based traditional systems and the DJSCC-based SC system over the AWGN channel under different BCR constraints and SNRs. We use the optimized VideoQA model ($g(\cdot;\boldsymbol{\zeta})$ and $u(\cdot;\boldsymbol{\nu})$) with lossless video and text to execute VideoQA tasks as the upper bound performance of the VideoQA-SC. The lower bound performance of the VideoQA-SC is obtained by performing VideoQA tasks only using texts at the receiver without video transmission. For fair comparison, the same VideoQA model optimized in training stage 1 is adopted in all comparison SC schemes. Content-adaptive and SNR-adaptive VideoQA-SCs are trained and tested under fixed SNRs and mixed SNRs, respectively.

\begin{figure}[!t]
\centering
\includegraphics[width=0.4\textwidth]{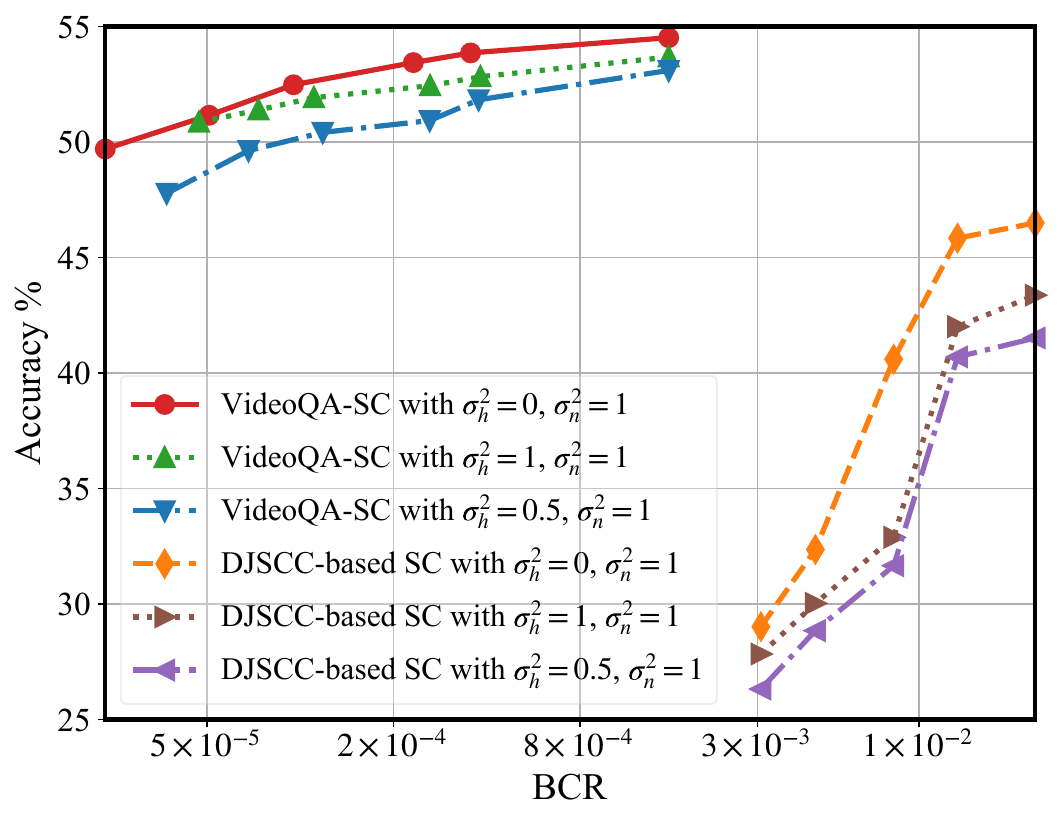}
\caption{Answer accuracy of VideoQA-SC over Rayleigh fading channels with different fading parameters $\sigma_{h}$.}
\label{Fig:fading}
\end{figure}

By comparing Fig. \ref{Fig:Action_All}\subref{Fig:A_0dB}, \subref{Fig:A_5dB} and \subref{Fig:A_10dB} or Fig. \ref{Fig:Transition_All}\subref{Fig:T_0dB}, \subref{Fig:T_5dB} and \subref{Fig:T_10dB}, it can be found that the proposed VideoQA-SC outperforms other SC schemes especially in low SNRs ($0$ dB). Since comparison SC systems need to restore the original video at the pixel level,   When the BCR constraint $R<R_\text{min}$, comparison SC systems lack the necessary information to reconstruct the original videos, which leads to a rapid deterioration of VideoQA performance. As the SNR decreases, $R_\text{min}$ gradually increases, which indicates that more bandwidths are required to guarantee the basic performance of SC systems when the channel conditions are worse. The DJSCC-based video transmission model DVST also employs a bandwidth-adaptive approach and achieves excellent video reconstruction performance, e.g., peak signal-to-noise ratio with the same bandwidth. However, there are slight bandwidth savings for DJSCC-based SC systems compared with SSCC-based traditional systems for VideoQA tasks. It indicates that the adopted entropy-based bandwidth allocation limited to pixel-level video reconstruction is not effective for SC systems oriented to a particular intelligent task. Besides, although DJSCC-based SC systems can avoid the cliff effect and are robust to channel noise, they still face the risk of system collapse when $R<R_\text{min}$ cannot be satisfied.

\begin{figure*}[!t]
\centering
\includegraphics[width=0.9\textwidth]{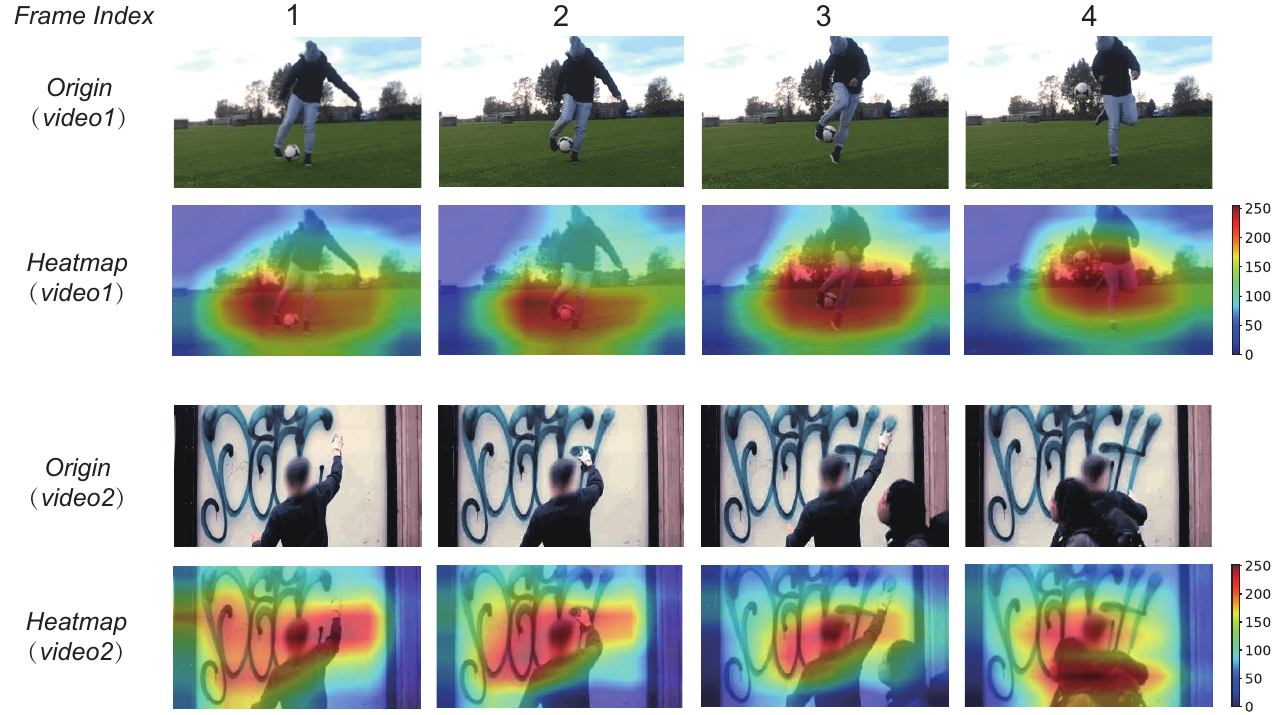}
\caption{Visualization of spatiotemporal correlations extracted by the semantic encoder.}
\label{Fig:relation}
\end{figure*}

Different from comparison schemes at pixel-level, the proposed VideoQA-SC extracts and processes video semantics at the object and frame levels, and directly performs VideoQA tasks based on the reconstructed video semantics at the receiver, which results in significant bandwidth savings with guaranteed VideoQA performance. In this way, the minimum BCR $R_\text{min}$ of VideoQA-SC can be very low, which ensures the basic performance of VideoQA-SC under a wide range of BCR constraints. It can be seen that both types of VideoQA-SC achieve accuracy of over $53\%$ and $64\%$ under all testing SNR and BCR constraints in the TGIF-QA-R Action and Transition dataset, respectively. Benefiting from E2E joint training and SNR-adaptive bandwidth allocation, the proposed VideoQA-SC demonstrates robustness to noisy wireless channels while approaching the upper bound performance. In Fig. \ref{Fig:Action_All}\subref{Fig:A_0dB}, the accuracy of SNR-adaptive VideoQA-SC exceeds that of the DJSCC-based SC system by $5.17\%$ with almost $0.05\%$ of its bandwidth.

We further show the performance of VideoQA-SC over Rayleigh block fading channels with different fading parameters $\sigma_{h}$ and fixed $\sigma_{n}^{2}=1$. In order to fix $\sigma_{n}^{2}$, we adjust the signal power to control the ratio of it to the simulated average noise power is $1$. If the perfect channel state information can be obtained at the transmitter, the statistical SNR and the current SNR are fed into rate predictors to improve the robustness of SNR-adaptive VideoQA-SC to Rayleigh block fading channels. Fig. \ref{Fig:fading} shows that VideoQA-SC is able to achieve slow and smooth degradation of VideoQA performance with the increasing fading, however, DJSCC-based SC fails to resist noise in Rayleigh block fading channels.

\begin{figure*}[!t]
\centering
\includegraphics[width=\textwidth]{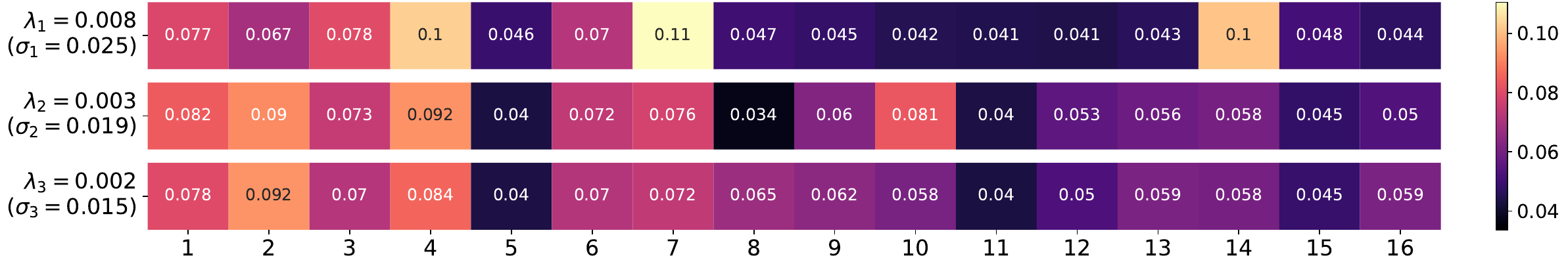}
\caption{Visualization of different initial values of rate embeddings. Each row denotes trained initial values of rate embeddings $\mathbf{Y}_\text{rate}$ under corresponding $\lambda$ (higher $\lambda$ corresponds to lower BCR $R$). The horizontal axis indicates the frame index and the initial values are normalized over all frames. $\sigma$ indicates the standard deviation of corresponding rate embeddings.}
\label{Fig:embedding}
\end{figure*}

\begin{figure*}[!t]
\centering
\includegraphics[width=\textwidth]{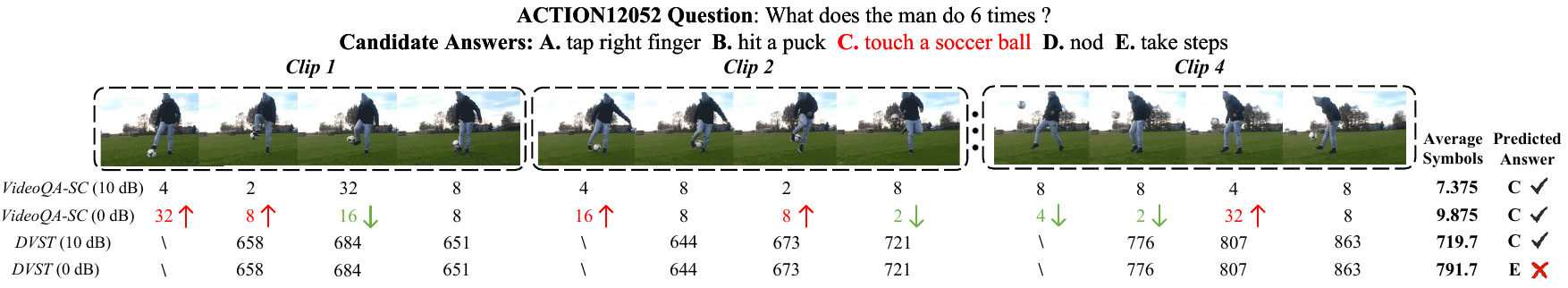}
\caption{An example of VideoQA-SC robust to the channel noise. Each row represents the bandwidth allocation made by the corresponding model for all video frames over the AWGN channel with a specific SNR. Each column denotes the number of complex transmission symbols allocated for the corresponding frame in different cases. The correct answer is marked in red.}
\label{Fig:visual}
\end{figure*}

\subsection{Visualization and Analysis}

We visualize the extracted semantics $\mathbf{Y}_v$ and the rate embeddings $\mathbf{Y}_\text{rate}$ trained under different $\lambda$ values to further analyze the operational mechanism of VideoQA-SC. Additionally, we conduct further experiments to demonstrate the ability of VideoQA-SC to handle different types of questions.

As shown in Fig. \ref{Fig:relation}, we first visualize the video semantics $\mathbf{Y}_{v}$ extracted by the semantic encoder to observe how it models the spatiotemporal correlations of videos. Given that the semantic encoder extract spatiotemporal correlations within one clip, we visualize one clip including four frames from each of two different videos. We employ the Grad-CAM \cite{Grad_CAM} method to obtain heatmaps of $\mathbf{Y}_{v}$ by overlaying the gradients obtained from backpropagation onto the original frames.

By comparing the frames in the first video, we can observe that the pixels near the football and the man's legs exhibit higher gradients, indicated by warmer colors. This demonstrates that the spatiotemporal extraction pay more attention to the interaction between the football and the man, specifically the action of ``kicking the football". Furthermore, the visualization of the second video illustrates the extraction of spatiotemporal correlations when additional objects suddenly appear in some frames. For the first two frames, the gradients are concentrated near the man's hand and the wall, suggesting that the extraction focus on the action of ``graffiti". However, when a bystander appears, the extraction pay attention to the interaction between the bystander and the man gradually, as the specific question is unknown. This ensures that the extracted video semantics are as comprehensive as possible and can be used to answer various questions.

As shown in Fig. \ref{Fig:embedding}, we then visualize the values of the trained initial values of rate embeddings $\mathbf{Y}_{\text{rate}}$ under three $\lambda$ values to explore their effect on the model. Given that the video semantics $\mathbf{Y}_{v}$ include 16 frames in our implementation, the initial values of $\mathbf{Y}_{\text{rate}}$ are normalized over all 16 frames, indicating the attention allocation to different frames. We can observe that the initial values corresponding to $\lambda_{1}$ exhibit extreme values in specific frames, which can be reflected in a higher $\sigma$. It demonstrates that when the bandwidth is limited, the network tends to perform VideoQA by some specific key frames. As $\lambda$ decreases from $\lambda_{1}$ to $\lambda_{3}$ ($R$ increases accordingly), the attention allocated to all frames approaches uniformity, which can be reflected in a lower $\sigma$. It indicates that when the available bandwidth is abundant, the model tends to utilize all frames equally to perform VideoQA.

Fig. \ref{Fig:visual} provides an example to show the robustness of VideoQA-SC to the AWGN channel. DVST is adopted as the comparison scheme for video transmission. In this case, every $4$ frames are combined into one clip for semantic extraction and video coding. The first frame is used as I-frame to give the reference of the coding of P-frames for DVST. 

When the channel SNR is $10$ dB, both VideoQA-SC and DVST successfully transmit the video and help to predict the correct answer. As the SNR drops to $0$ dB, DVST makes the same bandwidth allocation decision as when the SNR is $10$ dB, failing to predict the correct answer. In contrast, due to the utilization of estimated SNR as a prior to aid bandwidth allocation, VideoQA-SC is able to dynamically adjust the number of transmission symbols allocated to each frame, ultimately overcoming the channel noise and predicting the correct answer. Although VideoQA-SC allocates more average transmission symbols for each frame under a low SNR, it is still much lower than the number of transmission symbols allocated by DVST.

\begin{table}[!htbp]
    \begin{center}
    \renewcommand\tabcolsep{4.5pt}
    \begin{threeparttable}
    \caption{Performance of VideoQA-SC on NExT-QA dataset.}
    \label{tab1}
    \begin{tabular}{cccccc}
        \toprule
        \multicolumn{2}{c}{Model} & {Causal} & {Temporal} & {Descriptive} & {Average}\\
        \midrule
        \multicolumn{2}{c}{Unlimited Bandwidth} & 50.17\% & 53.03\% & 61.64\% & 52.83\% \\
        \midrule
        \multirow{2}{*}{$\lambda=0.001$} & Accuracy & -0.35\% & -2.82\% & -2.78\% & -1.41\% \\
        \cmidrule(){2-6}
        & BCR & 59.08 & 59.12 & 69.00 & 60.72 \\
        \midrule
        \multirow{2}{*}{$\lambda=0.0015$} &Accuracy & -1.10\% & -3.01\% & -2.92\% & -1.88\% \\
        \cmidrule(){2-6}
        & BCR & 24.00 & 24.68 & 29.04 & 25.04 \\
        \midrule
        \multirow{2}{*}{$\lambda=0.002$} &Accuracy & -1.37\% & -3.12\% & -4.49\% & -2.32\% \\
        \cmidrule(){2-6}
        & BCR & 7.44 & 7.61 & 8.76 & 7.71 \\
        \bottomrule
    \end{tabular}
    \begin{tablenotes}
            \footnotesize
            \item[1] For models trained with bandwidth constraints, average BCR are normalized by $10^{-6}$.
            \item[2] The training and testing are conducted under $\text{SNR}=10\text{dB}$.
    \end{tablenotes}
    \end{threeparttable}
    \end{center}
\end{table}

We further test VideoQA-SC on NExT-QA\cite{nextqa} dataset to provide a more detailed analysis of VideoQA tasks. Compared with TGIF-QA-R dataset, NExT-QA dataset including various question types (causal, temporal and descriptive) with higher video resolutions and longer durations. The results are shown in TABLE. \ref{tab1}. We use the trained VideoQA model with lossless video transmission (unlimited bandwidth) as the upper bound of comparison to measure the performance of VideoQA-SC under different bandwidth constraints. The accuracy changes of each question type relative to the upper bound are marked. In general, the accuracy of all three types of questions shows a slow decrease as the BCR decreases. One can be observed that compared to causal and temporal questions, the accuracy of descriptive questions is most affected by the reduction of BCR. This suggests that answering descriptive questions may heavily rely on the completeness of the video semantics. The experiments demonstrate that the accuracy of different types of questions are influenced differently by changes of transmission bandwidth. Overall, although we do not consider the specific designs for different question types, VideoQA-SC still achieves a good trade-off between bandwidth usage and task performance across a wide range of question types.

\section{Conclusion}\label{Section:Conclusion}
In this paper, we have proposed an E2E SC system, VideoQA-SC, to perform VideoQA tasks in wireless networks without relying on the reconstructed source. Taking advantage of the efficient video semantic extraction and the learning-based bandwidth-adaptive DJSCC transmission, VideoQA-SC is able to fully leverage video semantic information to improve system performance for VideoQA tasks with high bandwidth-efficiency and noise-robustness. VideoQA-SC is trained in an E2E manner with the goal of maximize the VideoQA performance under bandwidth constraints. Experiments show that the proposed VideoQA-SC outperforms SSCC-based traditional systems and advanced DJSCC-based SC systems under a wide range of channel conditions and bandwidth constraints. In particular, VideoQA-SC improves $5.17\%$ VideoQA accuracy while achieves nearly $99.5\%$ bandwidth savings compared with the DJSCC-based SC system over AWGN channel when SNR is $0$ dB, which demonstrates the great potential of SC system design for video applications. In this paper, we do not consider the specific designs for handling different question types, the accuracy of which are influenced differently by changes of transmission bandwidth. The future work will focus on the utilization of question types in semantic extraction and coding to make better trade-off between bandwidth usage and task performance across a wide range of question types.

\vfill

\end{document}